\documentclass{article}
\usepackage[table,xcdraw]{xcolor}
\usepackage{codefuse_tech_report}
\usepackage[colorlinks = true,
            linkcolor = blue,
            urlcolor  = blue,
            citecolor = blue,
            anchorcolor = blue]{hyperref}   
\usepackage{microtype}
\usepackage{hyperref}
\usepackage{xurl}
\usepackage{booktabs}
\usepackage{enumitem}
\usepackage{multicol}
\usepackage{CJKutf8}
\usepackage{amsmath}
\usepackage{siunitx}
\usepackage{floatflt}
\usepackage{graphicx}
\usepackage{booktabs}
\usepackage{authblk}
\usepackage{lipsum}
\usepackage{algorithm}
\usepackage{algorithmicx}
\usepackage{algpseudocode}
\usepackage{microtype}
\usepackage{subfigure}
\usepackage{multirow}
\usepackage{booktabs} 
\usepackage{amsmath}
\usepackage{amssymb}
\usepackage{graphicx}
\usepackage{enumitem}
\usepackage{caption}
\usepackage{subcaption}
\usepackage{booktabs} 
\usepackage{wrapfig}

\usepackage{pifont}  
\usepackage{bm}
\usepackage{ulem}
\usepackage{url}
\usepackage{threeparttable}
\usepackage{multirow}

\usepackage{adjustbox}

\def\1{\bm{1}}

\def\rvc{{\mathbf{c}}}

\def\rvh{{\mathbf{h}}}

\def\rvs{{\mathbf{s}}}

\def\vc{{\bm{c}}}

\def\vh{{\bm{h}}}

\def\vq{{\bm{q}}}

\def\mC{{\bm{C}}}

\def\mW{{\bm{W}}}

\DeclareMathAlphabet{\mathsfit}{\encodingdefault}{\sfdefault}{m}{sl}
\SetMathAlphabet{\mathsfit}{bold}{\encodingdefault}{\sfdefault}{bx}{n}
\newcommand{\tens}[1]{\bm{\mathsfit{#1}}}

\def\tK{{\tens{K}}}

\def\tQ{{\tens{Q}}}

\def\tV{{\tens{V}}}

\def\tX{{\tens{X}}}

\def\gA{{\mathcal{A}}}

\def\gD{{\mathcal{D}}}
\def\gE{{\mathcal{E}}}

\def\gO{{\mathcal{O}}}

\def\sR{{\mathbb{R}}}

\DeclareMathOperator{\vPMA}{vPMA}
\DeclareMathOperator{\LN}{LN}
\DeclareMathOperator{\MHA}{MHA}
\DeclareMathOperator{\FFN}{FFN}

\usepackage{hyperref}
\usepackage{amssymb} 

\algnewcommand{\LeftComment}[1]{\Statex \(\triangleright\) #1}

\usepackage{array}
\usepackage{amsmath}
\usepackage{amssymb}
\usepackage{mathtools}
\usepackage{amsthm}

\usepackage[capitalize,noabbrev]{cleveref}
\usepackage{adjustbox} 

\theoremstyle{plain}

\theoremstyle{definition}

\theoremstyle{remark}

\colmfinalcopy

\usepackage[utf8]{inputenc}
\usepackage[T1]{fontenc}
\usepackage{caption} 
\usepackage{adjustbox} 
\usepackage{arydshln}
\usepackage{fontawesome5}

\usepackage{tcolorbox}
\tcbuselibrary{skins,breakable}

\tcbuselibrary{skins}

\usepackage{color}
\usepackage{soul} 

\definecolor{tred}{RGB}{251, 130, 132}
\definecolor{torange}{RGB}{247, 162, 116}
\definecolor{tyellow}{RGB}{251, 218, 140}
\definecolor{tgreen}{RGB}{127, 204, 181}
\definecolor{tblue}{RGB}{89, 177, 215}
\definecolor{insightblue}{RGB}{162, 210, 255}
\definecolor{questionred}{RGB}{255, 175, 204}
\definecolor{customcolor}{RGB}{69, 148, 68} 

\newcommand{\lzh}[1]{\textcolor{black}{#1}}

\newcommand{\red}[1]{\textcolor{black}{#1}}
\newcommand{\rbt}[1]{\textcolor{black}{#1}}

\title{E2LLM: Encoder Elongated Large Language Models for Long-Context Understanding and Reasoning}

\author{%
Zihan Liao\thanks{Equal Contribution. This work was done when Zihan Liao was a research intern at Ant Group.}$^{\phantom{*},1,2}$
~~Jun Wang$^{*,2}$
~~Hang Yu$^{*,2}$
\\

\vspace{-10pt}
\bf
~~Lingxiao Wei$^{1}$ 
~~Jianguo Li\thanks{Correspondence to: Jianguo Li \textless lijg.zero@antgroup.com\textgreater, Jun Wang \textless wongjun@gmail.com\textgreater ~and Wei Zhang \textless zhangwei.thu2011@gmail.com\textgreater.}~~$^{2}$
~~Jun Wang$^{\dag,1}$
~~Wei Zhang$^{\dag,1}$

\vspace{10pt}

$^1$East China Normal University\ \ \ $^2$Ant Group\\
\vspace{10pt}
\hspace{-10pt}\faGithub ~\url{https://github.com/codefuse-ai/E2LLM}\\
}

\colmfinalcopy 

\begin{document}

\maketitle


\begin{abstract}
Processing long contexts is increasingly important for Large Language Models (LLMs) in tasks like multi-turn dialogues, code generation, and document summarization. This paper addresses the challenges of achieving high long-context performance, low computational complexity, and compatibility with pretrained models -- collectively termed the ``impossible triangle''. We introduce E2LLM (Encoder Elongated Large Language Models), a novel approach that effectively navigates this paradox. E2LLM divides long contexts into chunks, compresses each into soft prompts using a pretrained text encoder, and aligns these representations with a decoder-only LLM via an adapter. To enhance the LLM's reasoning with these soft prompts, we employ two training objectives: encoder output reconstruction and long-context instruction fine-tuning. Extensive experiments reveal that E2LLM not only outperforms 8 state-of-the-art (SOTA) methods in effectiveness and efficiency for document summarization and question answering, but also achieves the best performance on LongBench v2 among models of comparable size. 
\end{abstract}

\section{Introduction}
Understanding and reasoning about long context has become essential for LLMs, especially for tasks like multi-round dialogues~\citep{bai2024mt}, (multi)-repository code generation~\citep{zhang2023repocoder}, and (multi)-document summarization~\citep{giorgi2023open} and question answering~\citep{singh2021end}. These tasks often require processing thousands or even millions of tokens to ensure coherence and accuracy. In addition, techniques that effectively boost the performance of LLMs—such as chain-of-thought reasoning~\citep{wei2022chain}, in-context learning~\citep{dong2022survey}, and retrieving relevant documents or historical conversations~\citep{ding2024survey}—are also pushing the demand for longer context window.

\begin{wrapfigure}{r}{0.45\textwidth}
    \centering
    \includegraphics[width=\linewidth]{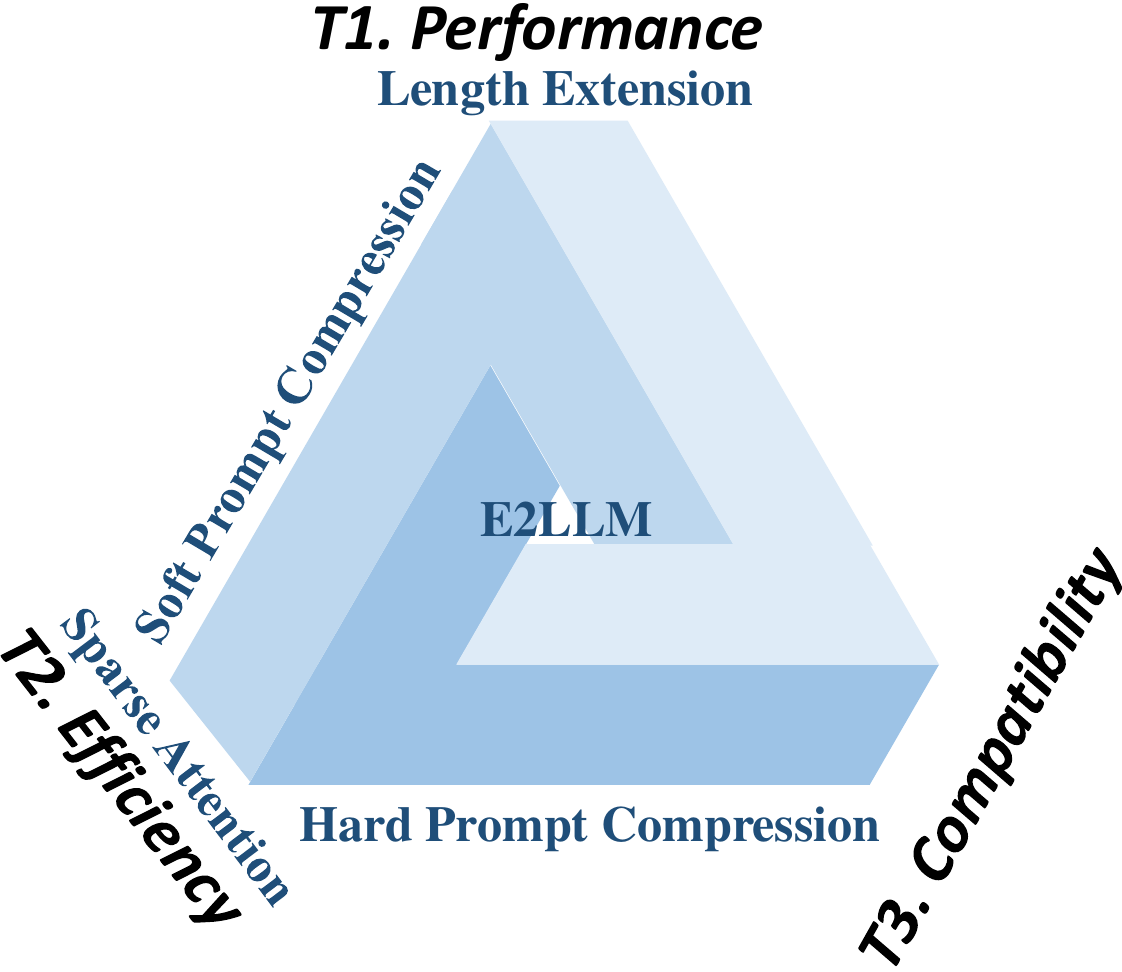}
    \caption{E2LLM solves the ``impossible triangle'' challenge of Performance, Efficiency, and Compatibility.}
    \label{fig:impossible_triangle}
    \vspace{-2pt}         
\end{wrapfigure}

Considerable efforts have been and are still being put into increasing the context length of LLMs, aiming at achieving strong performance for longer contexts (\textbf{\textit{T1}}), while reducing the training and inference complexity (\textbf{\textit{T2}}), and at the same time being compatible with pretrained models (\textbf{\textit{T3}}). Achieving this compatibility is crucial for effectively leveraging the pretrained knowledge contained in these models, allowing for \textbf{parameter and sample efficiency} without necessitating extensive additional training with large datasets. However, achieving all three targets simultaneously presents a formidable challenge that often leads to some compromises, a phenomenon we refer to as the ``impossible triangle'', as illustrated in Figure~\ref{fig:impossible_triangle}. Current research in this field primarily focuses on three main avenues: modifying position embeddings, attention mechanisms, and the long input sequence itself.\footnote{A detailed literature review is provided in Appendix~\ref{app:related_works}.}

\noindent\textbf{Length Extension}: The first group of methods adjust the position embeddings of LLMs to accommodate longer context extensions~\citep{peng2023YaRN,dinglongrope}. This typically involves selecting a large base value for RoPE~\citep{su2024roformer} followed by continued pretraining or fine-tuning~\citep{zhao2024longskywork,gao2024train}. While these methods effectively extend context length (\textbf{\textit{T1}}), they typically require substantial continued pretraining with tens of billions of tokens (\textbf{\textit{T3}}) and incur high computational complexity during both training and inference (\textbf{\textit{T2}}). For instance, even with the ability to extend context window to 2M, as seen in LongRoPE~\citep{dinglongrope}, enormous resources are required to train and deploy the model, and inference times can be prohibitively long for extended sequences. 

\noindent\textbf{Sparse Attention}: The second group replaces full attention in LLMs with local attention or a combination of global and local attention~\citep{chen2023extending,xiao2023efficient}. This approach significantly reduces the quadratic complexity associated with full attention, even achieving linear complexity in theory (\textbf{\textit{T2}}). However, a notable concern with sparse attention is its potential to neglect informative history, as certain tokens may not be attended to during the attention calculations (\textbf{\textit{T1}}). Moreover, since LLMs are not originally pretrained with sparse attention, adapting them to sparse attention may require extensive training or fine-tuning (\textbf{\textit{T3}}). 

\noindent\textbf{Prompt Compression}: The third group of strategies directly compresses the input sequence to reduce its length (\textbf{\textit{T2}}), which can be further divided into two subcategories. The first subgroup, known as \textbf{hard prompt compression}—exemplified by methods such as Retrieval-Augmented Generation (RAG)~\citep{ding2024survey} and LLMLingua~\citep{jiang2023llmlingua}—tends to process compression and inference in a two-step manner. As a result, any loss of information or introduction of irrelevant content during the compression stage may adversely affect performance in the subsequent inference step (\textbf{\textit{T1}}). Alternatively, the second subgroup considers \textbf{soft prompt compression}, which summarizes long contexts into embedding vectors~\citep{chevalier2023adapting,tan2024lloco}. However, utilizing LLMs in these approaches to directly generate sentence-level embeddings diverges from their original pretraining objective of next token prediction. Consequently, achieving satisfactory performance in this context often demands rigorous training or fine-tuning to align the model's capabilities with the new objective (\textbf{\textit{T3}}).

\begin{figure*}[!t]
\vspace{-3ex}
    \centering
    \includegraphics[width=\linewidth]{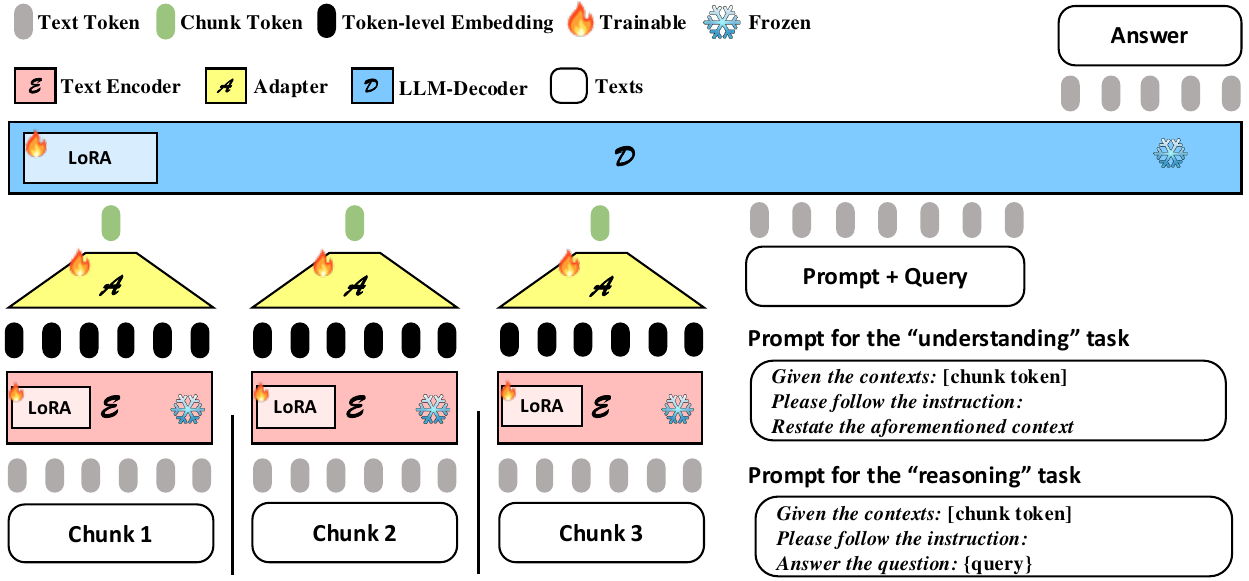}
    \vspace{-4ex}
    \caption{The E2LLM architecture.}
    \label{fig:arch}
    \vspace{-2ex}
\end{figure*}

In this paper, we propose a novel compression based method named E2LLM (\textbf{E}ncoder \textbf{E}longated \textbf{L}arge \textbf{L}anguage \textbf{M}odels) that adeptly navigates the complexities of the ``impossible triangle''. As shown in Figure~\ref{fig:arch}, we first divide a long context into smaller chunks. A pre-trained text encoder (e.g., BERT) then processes each chunk, generating embeddings for the tokens within it. An adapter subsequently aggregates these token-level embeddings into a single, representative ``chunk token'' (a chunk-level embedding) and projects it into the input embedding space of the LLM decoder. To ensure alignment between the encoder and decoder, we employ two training objectives: reconstruction of the original input text from the encoded embeddings (``understanding'') and long-context instruction fine-tuning (``reasoning''). 

We postulate that \textbf{LLMs are inherently rich in knowledge}; thus, \textbf{properly compressed soft prompts (i.e., the chunk tokens) can succinctly convey adequate information for LLMs to generate accurate answers}. Moreover, \textbf{pre-trained encoder models are inherently crafted to produce chunk-level representations}. As a result, this design allows E2LLM to leverage the strengths of both pre-trained encoders and decoders, minimizing the need for extensive additional training (\textbf{\textit{T3}}). 
Additionally, compressing each original chunk into a single vector (the chunk token) not only enhances training and inference efficiency (\textbf{\textit{T2}}) but also scales up the context length significantly (\textbf{\textit{T1}}). Indeed, the theoretical context window equals the product of the encoder and decoder sequence lengths.
Experiments show that E2LLM outperforms 8 baselines in long document summarization and QA using the same pretrained LLM and limited fine-tuning data. It also achieves \textbf{the best performance} on the challenging \textbf{LongBench v2}~\citep{bai2024longbench2} benchmark \textbf{among comparably sized models}. These results validate E2LLM’s effectiveness in balancing performance, efficiency, and compatibility. To summarize, our main contributions are:

\begin{itemize}[leftmargin=*, itemsep= 0pt, topsep = 0pt, partopsep=0pt]
\item We propose E2LLM, a novel long-context modeling framework built on pre-trained text encoders and decoder-only LLMs to effectively address the ``impossible triangle'' challenge.


\item We introduce two training objectives: soft prompt reconstruction and long-context instruction fine-tuning, enabling the LLM to understand the soft prompt while reasoning about accurate outputs. 

\item Comprehensive experiments conducted on diverse benchmarks demonstrate the efficiency and practicality of E2LLM and reveal its superiority over 8 SOTA baselines and competency on LongBench v2.
\end{itemize}

\section{Our Approach: E2LLM}
In this section, we detail E2LLM, which effectively combines the strengths of pre-trained text encoders and LLM decoders to facilitate understanding and reasoning over long contexts.

\subsection{Model Architecture}
\label{sec:model_arch}
Figure~\ref{fig:arch} illustrates the architecture of the E2LLM framework, which comprises four key components: a Chunker, a Text Encoder $\gE_\theta$, an Adapter $\gA_\phi$, and an LLM Decoder $\gD_\eta$. Here, $\theta$, $\phi$, and $\eta$ denote the (learnable) parameters specific to each component. For long input contexts, E2LLM first performs chunking. Each resulting chunk is processed by the encoder, which captures its token-level representations (the black tokens in Figure~\ref{fig:arch}). The adapter then aggregates these token-level embeddings into a single representative ``chunk token'' (a chunk-level embedding; see the green tokens in Figure~\ref{fig:arch}) and maps it to the input embedding space of the LLM decoder, allowing the decoder to interpret the chunk effectively. Ultimately, the decoder utilizes these embeddings as substitutes for the original context and executes two fine-tuning tasks—“understanding” and “reasoning”—to train the entire framework. It is essential to note that \textbf{the choice of models for the encoder and decoder, the method of chunking, and the network architecture of the adapter can be customized to meet the needs of different domains.} E2LLM serves as a flexible framework, seamlessly integrating these components to effectively manage long contexts while \textbf{being capable of leveraging the power of more advanced components when available}. We will now introduce each component in detail. 

\vspace{0.5ex}
\noindent\textbf{Chunker:} The Chunker is responsible for dividing long contexts into smaller, manageable chunks while ensuring that the token length of each chunk does not exceed the maximum sequence length of the text encoder. Here, we adopt a straightforward yet effective approach: first, we define a chunk size and extract the initial chunk, then backtrack within this chunk to find breakpoints, such as periods or line breaks. After locating these breakpoints, a new chunk begins at the end of the previous one, and the backtracking method is applied again. We repeat this process until the entire text is chunked. This method helps maintain the semantic integrity of the original texts. Other methods, such as introducing overlap between chunks, can also benefit E2LLM. Our experiments in Appendix~\ref{app:hyperparameter} indicate that chunk size can influence performance. Including excessive context within a single chunk can degrade performance because a high compression ratio may render the embedding vector too generic, compromising specificity. Conversely, using excessively small chunk sizes can disrupt the semantic integrity of sentences, negatively impacting performance. Furthermore, we highlight that the \textbf{impact of the chunker in E2LLM is less pronounced when aligning the encoder and decoder}, as introduced in the sequel. In contrast to RAG, where the retriever (encoder) and the generator (decoder) are two distinct models without alignment, E2LLM benefits from this cohesion. \textbf{This alignment minimizes the risk of inconsistency in text interpretation}, which can arise when models are pretrained on different corpora and objectives. More discussions on chunk size is provided in Appendix~\ref{app:hyperparameter}.

\vspace{0.5ex}
\noindent\textbf{Text Encoder $\gE$:} After chunking, each chunk is input into the text encoder. Pre-trained encoders, such as GTE~\citep{li2023generaltextembeddingsmultistage} and BGE~\citep{xiao2023c}, are typically trained using contrastive learning, allowing the [CLS] token to serve as the embedding vector. However, this token primarily captures the discriminative features necessary for differentiating between chunks and often neglects valuable information that could benefit the LLM decoder when generating responses. To address this limitation, we do not rely on the [CLS] token as the direct chunk embedding. Instead, we extract the token-level embeddings produced by the encoder and pass them into a subsequent adapter for aggregation. Additionally, we implement low-rank adaptation (LoRA)~\citep{hu2021lora} to enable fine-tuning of the text encoder during the alignment process. These strategies enhance the encoder's ability to retain and extract more relevant information from the original text in each chunk, ultimately improving the performance of the LLM.

\vspace{0.5ex}
\noindent\textbf{Adapter $\gA$:} To facilitate the LLM's understanding of the chunk-wise semantics derived from the encoder's output, we introduce an adapter that serves two primary functions: (i) to compress the encoder's outputs into a single chunk token, and (ii) to project this token into the input embedding space of the LLM. To achieve these objectives, we propose a variant of Pooling by Multihead Attention (vPMA)~\citep{lee2019set}, that is,
\begin{align}
    \vc &= \vPMA(\tX) = \LN(\vh + \FFN(\vh)),  \label{eq:pma}\\
    \vh &= \LN(\MHA(\vq, \tK, \tV) + \vq), \\
    \tK &= \tX\mW_K,\quad \tV = \tX\mW_V,
\end{align}
where $\LN$, $\FFN$, and $\MHA$ respectively denotes layer normalization, row-wise feedforward network, and multihead attention. In this formulation, $\tX \in \sR^{C \times d_\text{enc}}$ represents the token-level embeddings with length $C$ and dimension $d_\text{enc}$ given by the text encoder for a specific chunk, $\vq \in \sR^{1 \times d_\text{dec}}$ represents the learnable query or seed vector with length $1$ and dimension $d_\text{dec}$, and $\mW_K, \mW_V \in \sR^{d_\text{enc}\times d_\text{dec}}$ are the projection matrices. 

By computing the multihead attention $\MHA$ between $\vq$ and $\tK$, we simultaneously compress all token-level embeddings into a single chunk token $\vc\in\sR^{1 \times d_\text{dec}}$, while also mapping the dimensions from the encoder ($d_\text{enc}$) to the decoder ($d_\text{dec}$). The resulting token $\vc$ for this chunk is referred to as the \textbf{chunk token} or \textbf{soft prompt}. Note that the original PMA does not alter the dimensions, so we adjust $\vq$ to ensure effective alignment between the encoder and decoder dimensions. 

\vspace{0.5ex}
\noindent\textbf{LLM Decoder $\gD$:} Finally, we concatenate the chunk tokens (the green tokens in Figure~\ref{fig:arch}) and the text tokens corresponding to the prompt and query, and ask the LLM to generate the answer for the query. Additionally, we employ LoRA to fine-tune the Decoder as part of the alignment process between the encoder and decoder.


\subsection{Training Tasks}
\label{ssec:train_tasks}
Now we focus on training the the adapter as well as the LoRA branch of the encoder and the decoder to enhance the E2LLM’s ability to comprehend lengthy input contexts and effectively reason about the corresponding answers. To accomplish this, we introduce two distinct training tasks.

\vspace{0.5ex}
\noindent\textbf{Understanding:}
The first task is designed to improve the LLM's understanding of the input. As depicted in Figure~\ref{fig:arch}, once the LLM receives chunk tokens from the adapter, we prompt it to restate or reconstruct the input. We refer to this as the ``understanding'' task. The specific prompt used is ``Given the contexts: [chunk token]\texttt{\textbackslash n} 
Please follow the instruction:\texttt{\textbackslash n}Restate the aforementioned context''. 

Notably, this task is self-supervised, allowing us to curate a significant amount of training data to ensure that the LLM comprehensively grasps the embeddings provided by the adapter. However, in our experiments, we utilize only the input from long-context instruction fine-tuning data for this task. Given that these inputs are often too lengthy to be fully reconstructed at once, we employ a sliding window approach, reconstructing the original context in segments based on a few consecutive chunks until the entire input has been restated.

\vspace{0.5ex}
\noindent\textbf{Reasoning:}
On the other hand, the second training task enables the LLM to generate answers based on the chunk tokens (i.e., the long context) and the user's query. We refer to this as the “reasoning” task, and the prompt crafted for this purpose is ``Given the contexts: [chunk token]\texttt{\textbackslash n} 
Please follow the instruction: \texttt{\textbackslash n} 
Answer the question: \{query\}''.

It is important to note that the ``understanding'' task serves as an auxiliary task, while our primary focus remains on the “reasoning” task. We determine the final checkpoints exclusively based on the validation loss associated with the ``reasoning'' task. In this context, we do not anticipate that E2LLM can achieve lossless compression of the context. However, we believe that the LLM decoder is capable of retaining or comprehending essential information from the context. The LLM operates as a “suggestion feature” for input methods, leveraging hints to generate meaningful responses. In this case, the chunk tokens provided by the text encoder serve as these essential hints. 

\subsection{Remarks}
\noindent\textbf{Relations to Other Works:} \lzh{For a detailed discussion on the relationship between E2LLM and other methods such as vision-language models (VLMs), retrieval-augmented generation (RAG), and Transformers-based architectures, please refer to Appendix~\ref{sec:relation_others}.}


\vspace{0.5ex}
\noindent\textbf{Maximum Context Window:} Theoretically, the maximum sequence length of E2LLM equals the product of the encoder and decoder's sequence lengths. However, as previously mentioned, setting the chunk size to match the encoder's sequence length presents challenges, as it may hinder the encoder's ability to retain all pertinent information within a single chunk. Thus, we need to choose a proper chunk size. As a result, the practical length of E2LLM is determined to be the chunk size multiplied by the sequence length of the LLM's decoder. In actuality, we set the maximum chunk size of 512 characters, which is approximately equivalent to 100 tokens. Hence, the context length has been expanded by nearly 100 times. When using Llama2-7B-chat as the decoder with a sequence length of 4K tokens, the final context window of E2LLM reaches approximately 400K.

\vspace{0.5ex}
\noindent\textbf{Time and Space Complexity during Inference:} Let us denote the original context length (excluding the prompt or instruction) as $L$ and the chunk size in E2LLM as $C$. Therefore, the total number of chunks becomes $L/C$. For each chunk, the resulting time and space complexity from the text encoder is $\gO(C^2)$. Given that there are $L/C$ chunks, the overall complexity for the encoding step is $\gO(CL)$. In practice, since all chunks can be processed in parallel, the time complexity can be further reduced by a constant factor. Subsequently, we pass the $L/C$ chunk tokens to the LLM decoder, which yields a complexity of $\gO(L^2/C^2)$. In summary, the total time and space complexity is $\gO(LC + L^2/C^2)$. To substantiate the efficiency of E2LLM during inference, we conduct empirical experiments that assess both inference time and memory usage (cf. Section \ref{sec:efficiency}). Moreover, we provide a discussion on the complexity of existing SOTA methods in Appendix~\ref{app:complexity}.

\section{Experiments}\label{exp}
In this section, we present a comprehensive evaluation of E2LLM. We begin by assessing its performance on two fundamental tasks: document summarization and QA. We then extend our evaluation to the recently proposed LongBench v2, a challenging benchmark that tests deep understanding and reasoning capabilities across real-world multitask scenarios. Additionally, we explore whether integrating retrieval mechanisms with E2LLM can further enhance its performance across different tasks. Finally, we analyze E2LLM's training and inference efficiency and conduct ablation studies to gain deeper insights into its components. Supplementary results on additional benchmark, Ruler, can be found in Appendix~\ref{app:ruler}.


\begin{table*}[t]
\vspace{-1ex}
\tabcolsep = 0.04cm
\begin{center}
\caption{Performance on Long-Context datasets. The best results are in \textbf{bold}, the second are \underline{underlined}, and the third are \uwave{wavy underlined}.}
\vspace{-1.5ex}
\label{table:performance_compare}
\begin{threeparttable}
\resizebox{\linewidth}{!}{
\renewcommand{\arraystretch}{1.3}
\begin{tabular}{lccccccccccccc}
\toprule
\multirow{2}{*}{Methods} &
  \multirow{2}{*}{\begin{tabular}[c]{@{}c@{}}Trainable\\ Parameters\end{tabular}} &
  \multirow{2}{*}{\begin{tabular}[c]{@{}c@{}}Context\\ Window\end{tabular}} &
  \multirow{2}{*}{\begin{tabular}[c]{@{}c@{}}Extension\\ Method\end{tabular}} &
  \multicolumn{2}{c}{QMSum} &
  \multicolumn{2}{c}{GovReport} &
  \multicolumn{2}{c}{Quality} &
  \multicolumn{2}{c}{NarrativeQA} &
  \multicolumn{2}{c}{TriviaQA} \\ \cline{5-14} 
 &
   &
   &
   &
  G-mean$\uparrow$ &
  PPL$\downarrow$ &
  G-mean$\uparrow$ &
  PPL$\downarrow$ &
  F1$\uparrow$ &
  PPL$\downarrow$ &
  F1$\uparrow$ &
  PPL$\downarrow$ &
  F1$\uparrow$ &
  PPL$\downarrow$ \\ \toprule
Llama2-7B-chat &
  0M &
  4K &
  - &
  11.51 &
  84.92 &
  5.50 &
  9.04 &
  9.38 &
  1,688.10 &
  4.65 &
  2,111.23 &
  12.06 &
  1,956.51 \\
StreamingLLM &
  0M &
  4M &
  Sparse Attn. &
  3.62 &
  220.12 &
  4.51 &
  330.54 &
  2.00 &
  230.72 &
  OOM &
  OOM &
  14.53 &
  596.87 \\
LongLoRA &
  140M &
  100K &
  Sparse Attn. &
  8.98 &
  \uline{14.48} &
  \uline{16.35} &
  \uline{2.88} &
  7.65 &
  381.32 &
  OOM &
  OOM &
  19.69 &
  438.25 \\
CEPE &
  1.31B &
  128k &
  Len. Exten. &
  10.77 &
  154.16 &
  4.82 &
  52.32 &
  2.33 &
  1,192.35 &
  OOM &
  OOM &
  - &
  - \\
YaRN &
  17M &
  64K &
  Len. Exten. &
  \uwave{12.31} &
  \uwave{16.22} &
  6.72 &
  \uwave{2.94} &
  \uline{13.80} &
  \uwave{31.32} &
  OOM &
  OOM &
  \uwave{20.22} &
  106.43 \\
RAG &
  0M &
  $+\infty$ &
  Hard Comp. &
  12.17 &
  17.21 &
  \uwave{14.45} &
  5.88 &
  4.85 &
  125.49 &
  \uwave{5.03} &
  48.42 &
  11.15 &
  98.96 \\
LongLLMLingua &
  0M &
  40K &
  Hard Comp. &
  8.93 &
  17.55 &
  4.56 &
  23.53 &
  10.89 &
  51.91 &
  4.53 &
  \uwave{31.36} &
  14.01 &
  \uwave{76.06} \\
LLoCO &
  17M &
  128K &
  Soft Comp. &
  \uline{12.99} &
  46.32 &
  5.73 &
  6.42 &
  \textbf{14.37} &
  \uline{9.44} &
  \uline{10.87} &
  \uline{16.88} &
  \textbf{63.21} &
  \uline{10.80} \\ \midrule
E2LLM &
  11M &
  400K &
  Soft Comp. &
  \textbf{15.47} &
  \textbf{13.66} &
  \textbf{18.43} &
  \textbf{2.81} &
  \uwave{12.95} &
  \textbf{8.99} &
  \textbf{12.47} &
  \textbf{13.07} &
  \uline{38.57} &
  \textbf{7.53} \\ \bottomrule
\end{tabular}
}

\begin{tablenotes}
    \scriptsize
    \item[*] For complete experimental results with more metrics, please refer to Table~\ref{table:performance_more_metrics} in Appendix~\ref{app:doc_sum_qa_discuss}.
    \end{tablenotes}
\end{threeparttable}
\end{center}
\vspace{-4ex}
\end{table*}

\vspace{-1ex}
\subsection{Document summarization and QA}
\label{ssec:doc_sum_qa}

We begin by evaluating E2LLM's performance on two fundamental long-context tasks: document summarization and QA. For summarization, we utilize QMSum and GovReport datasets, while for QA, we employ Quality, NarrativeQA, and TriviaQA datasets. Detailed dataset characteristics can be found in Appendix~\ref{app:doc_sum_qa_data} and Table~\ref{table:dataset_statistics}. Notably, Quality and TriviaQA have relatively shorter lengths compared to the summarization datasets, while NarrativeQA is significantly longer.

For comparison, we benchmark E2LLM against 8 baselines. These include \textbf{length extension} techniques such as YaRN~\citep{peng2023YaRN} and CEPE~\citep{yen2024long}, \textbf{sparse attention} strategies like StreamingLLM~\citep{xiao2023efficient} and LongLoRA~\citep{chen2023longlora}, as well as \textbf{hard and soft prompt compression} methods including RAG~\citep{gao2024retrievalaugmentedgenerationlargelanguage}, LongLLMLingua~\citep{jiang2023LongLLMLingua}, and LLoCO~\citep{tan2024lloco}. All baselines are built upon the same foundation model, Llama2-7B-chat, with CEPE and LLoCO having been continually pre-trained on this model using large corpora. We also include the original Llama2-7B-chat as an additional baseline. An overview of these methods is provided in Appendix~\ref{app:baselines}.

For our experiments, we train E2LLM and other training-dependent baselines \textbf{separately for each dataset}. We use the original validation sets for testing, while splitting the original training sets into training and validation subsets with a 95:5 ratio. For summarization, we use the \textbf{geometric mean (G-mean) of Rouge metrics}, which compare n-gram overlap between generated and reference texts. For QA, we calculate the \textbf{F1 score} based on unigram overlap between generated and reference answers. We also compute \textbf{perplexity (PPL)} of the correct answer across all datasets as a semantic-level metric to assess model prediction accuracy. Detailed metric descriptions are available in Appendix~\ref{app:doc_sum_qa_metrics}. The results are presented in Table~\ref{table:performance_compare}. Below, we analyze the performance of each group of methods\footnote{A more detailed discussion can be found in Appendix~\ref{app:doc_sum_qa_discuss}}:

\noindent\textbf{Soft prompt compression:} \textbf{E2LLM typically outperforms all methods}, across all datasets and metrics. While LLoCO also performs well in QA tasks, it shows limitations in summarization. This limitation stems from LLoCO's use of AutoCompressor~\citep{chevalier2023adapting}, where the chunk tokens only retain information necessary for subsequent chunks prediction. This approach, while effective for QA where selective context is sufficient, proves inadequate for summarization tasks requiring comprehensive context understanding. E2LLM overcomes this limitation by directly utilizing chunk tokens for various tasks rather than focusing on next-chunk prediction.

\noindent\textbf{Hard prompt compression:} Both LongLLMLingua and RAG underperform in Document summarization and QA tasks. Their drawbacks arise from (i) potential inconsistencies between retriever and generator interpretations and (ii) the selective use of document chunks. LongLLMLingua is further hampered by (i) insufficient token count for full chunk summarization and (ii) a non-consecutive token selection process that impairs LLM comprehension. E2LLM tackles these problems by aligning its encoder and decoder and utilizing all available document chunks as input.
\begin{wraptable}{l}{0.55\textwidth}
  \centering
  \caption{Performance comparison in terms of accuracy on LongBench v2 (w/o COT).}
  \label{table:performance_longbench_v2}
  \vspace{-1ex}          

  \begin{adjustbox}{width=\linewidth}
    \tabcolsep=1pt
    \renewcommand{\arraystretch}{1}
    \begin{tabular}{l|c|c|cc|ccc|c}
      \toprule
      Model & Size & Overall & Easy & Hard & Short & Medium & Long & Rank \\ \midrule
      o1-mini & - & 37.8 & 38.9 & 37.1 & 48.6 & 33.3 & 28.6 & 14 \\
      Mistral Large & 123B & 34.4 & 38.0 & 32.2 & 41.7 & 30.7 & 29.6 & 15 \\
      \rowcolor[HTML]{C0C0C0}E2LLM-7B & 7B & 31.8 & 33.3 & 30.9 & 37.8 & 28.4 & 28.7 & - \\
      Llama3.1-70B & 70B & 31.6 & 32.3 & 31.2 & 41.1 & 27.4 & 24.1 & 16 \\
      Nemotron-70B & 70B & 31.0 & 32.8 & 29.9 & 38.3 & 27.9 & 25.0 & 17 \\
      Qwen2.5-7B-ft & 7B & 31.2 & 31.3 & 31.2 & 39.4 & 27.0 & 25.9 & - \\
      NExtLong-8B & 8B & 30.8 & 33.9 & 28.9 & 37.8 & 27.4 & 25.9 & 18 \\
      GLM-4-9B & 9B & 30.2 & 30.7 & 29.9 & 33.9 & 29.8 & 25.0 & 19 \\
      Llama3.1-8B & 8B & 30.0 & 30.7 & 29.6 & 35.0 & 27.9 & 25.9 & 20 \\
      \rowcolor[HTML]{C0C0C0}Qwen2.5-7B & 7B & 30.0 & 30.7 & 29.6 & 40.6 & 24.2 & 24.1 & 21 \\
      Llama3.3-70B & 70B & 29.8 & 34.4 & 27.0 & 36.7 & 27.0 & 24.1 & 22 \\
      GPT-4o mini & - & 29.3 & 31.1 & 28.2 & 31.8 & 28.6 & 26.2 & 23 \\
      LLoCO & 7B & 28.2 & 30.2 & 26.0 & 36.8 & 24.2 & 21.5 & - \\
      Command R+ & 104B & 27.8 & 30.2 & 26.4 & 36.7 & 23.7 & 21.3 & 24 \\
      Mistral Large 2 & 123B & 26.6 & 29.7 & 24.8 & 37.8 & 19.5 & 22.2 & 25 \\ \bottomrule
    \end{tabular}
  \end{adjustbox}
\end{wraptable}

\noindent\textbf{Sparse attention:} LongLoRA shows strength in summarization but weakness in QA tasks, due to its shift shot attention mechanism. While beneficial for global information flow in summarization, the sparse attention mask limits information exchange between arbitrary tokens, potentially missing crucial context in QA tasks. StreamingLLM's training-free approach with a 
$\Lambda$-shaped attention mask further restricts information flow, leading to suboptimal performance. E2LLM maintains superior performance by using full attention with effective passage compression.

\noindent\textbf{Length extension:} YaRN achieves balanced performance across tasks but falls short of E2LLM due to attention dispersion (e.g., ``lost in the middle'') in long contexts~\citep{liu2024lost}. CEPE faces similar challenges, compounded by insufficient training data for its cross-attention layers (adapter) and architectural limitations in handling varying context lengths. E2LLM avoids these issues by not extending the decoder's original length but training it to effectively interpret encoder-encoded soft prompts.

\vspace{-1ex}
\subsection{LongBench v2}
\vspace{-0.5ex}
\label{ssec:lb_v2}
To further our method, we utilize the recently released LongBench v2 benchmark~\citep{bai2024longbench2}, recognized for its broad scope and difficulty. It consists of 503 multiple-choice questions with contexts ranging from 8K to 2M words, spanning 6 major tasks: single-document QA, multi-document QA, long in-context learning, long dialogue history understanding, code repository understanding, and long structured data understanding. Edited by nearly 100 professionals from diverse backgrounds, the benchmark's difficulty is evidenced by human experts achieving only 53.7\% accuracy within a 15-minute time limit per question. More details are available in Appendix~\ref{app:lb_v2}.

Instead of assessing the baseline models from the previous section, we compare E2LLM directly with the models on the leaderboard to demonstrate its practicality. Specifically, we utilize GTE-large-en as the encoder, and Qwen2.5-7B-Instruct as the decoder. We combine the training data used in the previous section, resulting in about 13,000 samples for encoder-decoder alignment. While the training data focuses on document QA and summarization, the benchmark encompasses a broader range of tasks. To ensure E2LLM's generalizability, we employ the SDFT method from~\citep{yang-etal-2024-self}, prompting the original decoder to generate answers for these training samples based on reference answers, thus mitigating the issue of catastrophic forgetting. The resulting model is referred to as E2LLM-7B. For comparison, we fine-tune the original Qwen2.5-7B-Instruct using the same training data, resulting in Qwen2.5-7B-ft. The results are summarized in Table~\ref{table:performance_longbench_v2}. Additionally, to demonstrate the efficiency of E2LLM, we compare E2LLM-7B with Qwen2.5-7B-ft in terms of inference time, as shown in Table~\ref{table:efficiency_longbench_v2}.

\begin{wraptable}{l}{0.5\textwidth} 
  \centering
  \vspace{-1ex}
    \tabcolsep=4pt
    \captionof{table}{Time cost (in seconds) comparison on LongBench v2.}
    \label{table:efficiency_longbench_v2}
    \vspace{-1ex}
    \resizebox{\linewidth}{!}{%
      \begin{tabular}{lcccc}
        \toprule
        Model & Framework & Short & Medium & Long \\ \midrule
        \multirow{2}{*}{Llama3.1-8B} & Transformers & 8.50s & OOM & OOM \\
                                     & vLLM         & 5.85s & 24.92s & 34.48s \\ \midrule
        \multirow{2}{*}{GLM4-9B}     & Transformers & 10.51s & OOM & OOM \\
                                     & vLLM         & 5.89s & 26.36s & 35.03s \\ \midrule
        \multirow{2}{*}{Qwen2.5-7B-ft} & Transformers & 7.13s & OOM & OOM \\
                                       & vLLM         & 5.75s & 24.67s & 34.31s \\ \midrule
        LLoCO-7B  & Transformers & 5.89s & 15.96s & 68.42s \\\midrule
        E2LLM-7B  & Transformers & 4.15s & 12.71s & 46.28s \\
        \bottomrule
      \end{tabular}}
\end{wraptable}
The results show that E2LLM demonstrates strong capabilities and efficiency in long-context modeling: (i) \textbf{Overall Performance:} E2LLM ranks 16th on the leaderboard, outperforming all similarly-sized models, including the recent NExtLong-8B~\citep{gao2025nextlong}. It achieves relative gains of 6.00\% and 1.92\% over Qwen2.5-7B-Instruct and Qwen2.5-7B-ft respectively, while surpassing larger models like Llama3.1-70B and Nemotron-70B. (ii) \textbf{Extreme Long-context Handling:} E2LLM-7B excels in the ``Long'' category (with 128K-2M words), \textbf{achieving a 19.09\% relative improvement over Qwen2.5-7B-Instruct and outperforming o1-mini}. (iii) \textbf{Inference Efficiency:} E2LLM reduces inference time by 41.8\% compared to original Qwen in the ``Short'' category when both use the Transformer framework. Notably, E2LLM remains faster than Qwen, even when the former uses the slower Transformer framework while the latter employs the faster vLLM, across ``short'' and ``medium'' datasets. In summary, by leveraging the strengths of pretrained encoder and decoder, \textbf{E2LLM achieves high accuracy and low inference time concurrently through parameter-efficient fine-tuning with just 13K samples}, effectively addressing the ``impossible triangle'' illustrated in Figure~\ref{fig:impossible_triangle}.

\begin{wraptable}{r}{0.5\textwidth} 
      \tabcolsep=0.3cm
    \renewcommand{\arraystretch}{1.}
    \captionof{table}{Performance gains resulting for different tasks from augmenting E2LLM with RAG.}
    \label{table:augmentation_with_retrieval}
    \resizebox{\linewidth}{!}{%
      \begin{tabular}{lcccc}
        \toprule
        Task Type & Doc. Sum. & Doc. QA & LB v2 & NIAH \\ \midrule
        Metric & G-mean & F1 & ACC & ACC \\ \midrule
        E2LLM-C   & 16.95 & 21.33 & 31.80 &  2.60 \\
        E2LLM-R   & 13.15 & 24.24 & 31.90 & 55.45 \\
        E2LLM-C+R & 16.49 & 22.70 & 31.80 & 55.84 \\ \bottomrule
      \end{tabular}}
\end{wraptable}

\vspace{-1ex}
\subsection{Augmentation with Retrieval}
\label{ssec:rag}

This section investigates the performance gains of augmenting E2LLM with RAG across various tasks. Our RAG implementation retrieves the top-$K$ most relevant chunks to a user query. The LLM decoder then receives both the text of these retrieved chunks and the chunk tokens representing the unselected chunks to generate a response. We evaluate this approach on document summarization and QA in Section~\ref{ssec:doc_sum_qa}, all tasks in LongBench v2 (which feature more complex QA than the document QA in Section~\ref{ssec:doc_sum_qa}), and the diverse Needle-in-a-Haystack (NIAH) tests within the Ruler benchmark~\citep{hsieh2404ruler}, designed to evaluate LLM recall capabilities. Performance metrics are averaged across datasets within each task. To assess the relative importance of chunk tokens and retrieved text, we compare E2LLM with retrieval (E2LLM-C+R) to the original E2LLM using only chunk tokens (E2LLM-C) and E2LLM relying solely on retrieved texts (E2LLM-R). Results are presented in Table~\ref{table:augmentation_with_retrieval}.

We observe that RAG significantly improves E2LLM performance on the NIAH task, likely due to the task’s need for fine-grained, token-level recall—a capability E2LLM-C, retaining only semantic-level chunk information, lacks. However, for summarization and QA, E2LLM-C+R performs comparably to E2LLM-C. In fact, E2LLM-C+R slightly degrades summarization performance, possibly due to noise introduced by retrieved text. Since summarization and QA are more common in real-world applications than synthetic NIAH tasks, E2LLM-C remains a practical and efficient tool for long-context modeling. For applications where fine-grained retrieval is paramount, augmenting E2LLM with RAG is a valuable solution.

\begin{figure}[t]
  \centering

  \begin{minipage}[b]{0.32\linewidth}
    \centering
    \includegraphics[width=\linewidth]{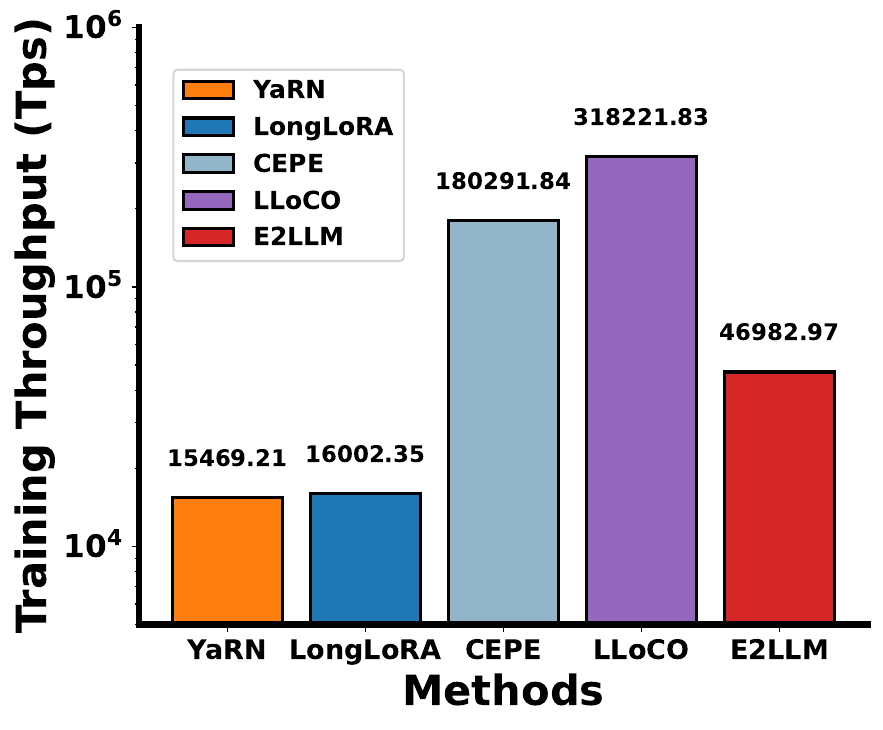}\\
    (a) Training Throughput (Tps)
    \label{fig:training_throughput}
  \end{minipage}
  \hfill
  \begin{minipage}[b]{0.32\linewidth}
    \centering
    \includegraphics[width=\linewidth]{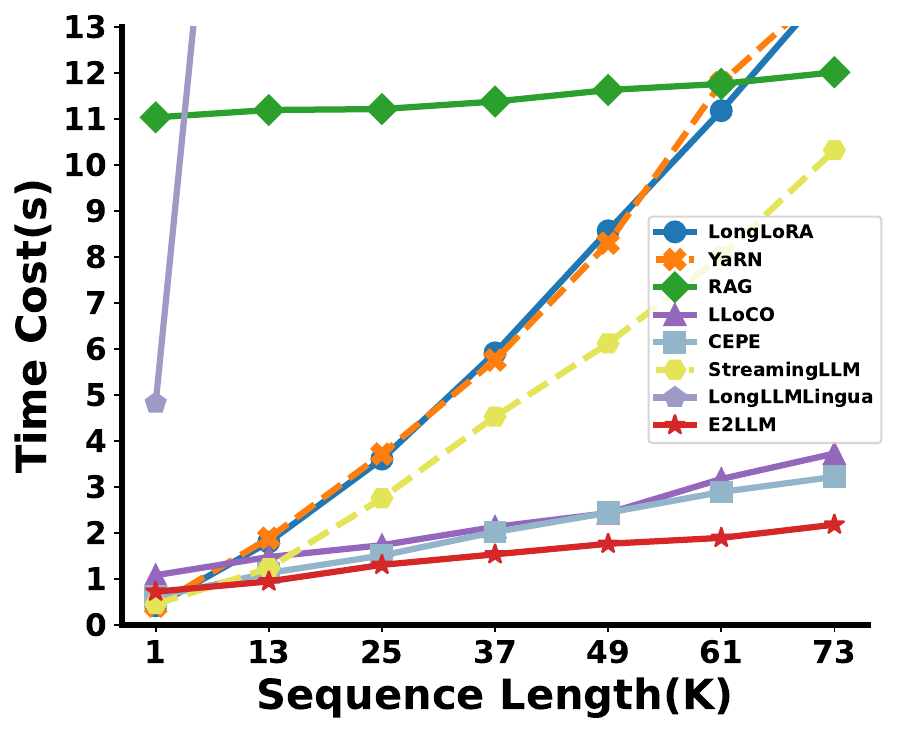}\\
    (b) Inference Time (s)
    \label{fig:time_usage}
  \end{minipage}
  \hfill
  \begin{minipage}[b]{0.32\linewidth}
    \centering
    \includegraphics[width=\linewidth]{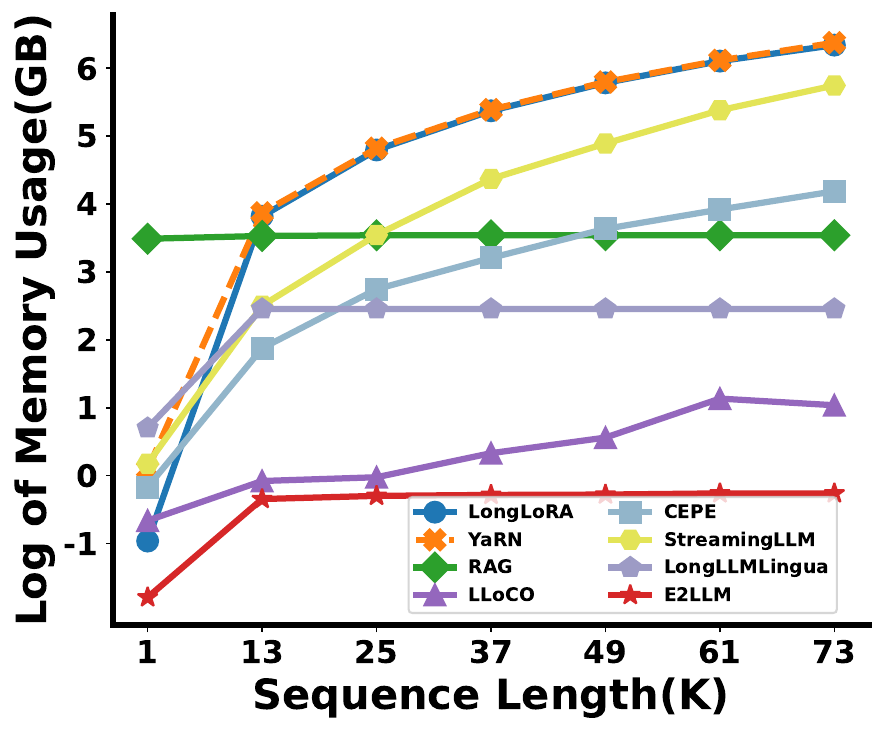}\\
    (c) Inference Memory (GB)
    \label{fig:memory_usage}
  \end{minipage}

  \vspace{0.5ex}
  \caption{Comparison of all methods on training and inference efficiency.}
  \label{fig:efficiency}
\end{figure}
\subsection{Training and Inference Efficiency}
\label{sec:efficiency}

We only present the conclusions here due to page limits; further discussions are in Appendix~\ref{app:efficiency}.

\vspace{0.5ex}
\noindent\textbf{Training Throughput:} Evaluation on eight A100 GPUs shows that CEPE, LLoCO, and E2LLM achieve significantly higher training throughput compared to YaRN and LongLoRA (Figure~\ref{fig:efficiency}(a)). LLoCO's efficiency stems from offline chunk token preparation, while CEPE benefits from linear scaling of its cross-attention layer with context length.

\vspace{0.5ex}
\noindent\textbf{Inference Time and Memory:} \textbf{E2LLM achieves the lowest runtime and memory usage, particularly for 73K-length sequences} (Figures~\ref{fig:efficiency}(b) and \ref{fig:efficiency}(c)). This efficiency is attributed to its high compression ratio (approximately 100x) compared to LLoCO's 32x, significantly reducing decoder token processing. Runtime behavior across all methods aligns with theoretical complexity (Table \ref{table:methods_time_space_complexity}).

\subsection{Ablation Study}
Again, we only provide the key findings here, with details in Appendix~\ref{app:ablation}. (i) Understanding'' loss and LoRA training for both encoder and decoder are essential. (ii) vPMA-based adapter outperforms MLP-based alternatives in chunk aggregation and encoder-decoder alignment. (iii) Incorporating overlap between chunks improves performance. (iv) \textbf{E2LLM benefits from stronger encoders and decoders}, implying that advances in open-source models can be leveraged. (v) We check E2LLM's sensitivity to hyperparameters, including the weight of the understanding’’ loss, the rank of LoRA for both encoder and decoder, and the head number in vPMA. Each factor has an optimal value. (vi) Varying chunk size has relatively small impact on performance (Table~\ref{table:chunk_size}), suggesting the alignment process in E2LLM effectively mitigates its influence. Still, selecting an optimal chunk size can lead to slight gains.

\vspace{-0.5ex}
\section{Conclusion}
\vspace{-1ex}
\lzh{In this paper, we present E2LLM, a novel approach to address the challenges of enhancing long-context performance in LLMs. It effectively navigates the ``impossible triangle'' by strategically splitting long contexts into chunks, compressing them into embedding vectors, and utilizing an adapter to align these representations with a decoder-only LLM. Two training objectives are employed to facilitate the understanding of soft prompts by the LLMs, resulting in superior performance in long-context scenarios. Experimental findings reveal that E2LLM effectively outperforms existing approaches in balancing the long-context performance, computational efficiency, and model compatibility. We believe that E2LLM offers a flexible framework for aligning text encoders and LLM decoders, with \textbf{considerable potential for enhancement as more powerful chunkers, encoders, adapters, and decoders become available.}}
\section{Acknowledgements}
We would like to thank Ant Group for supporting this work. This research was funded by Ant Group through the Ant Research Fund and, in part, by the National Natural Science Foundation of China (Grant No. 62572198).

\section*{Limitations}
Despite E2LLM's strong performance through data-efficient fine-tuning, collecting long-context fine-tuning data presents practical challenges. Thus, it is worthwhile to explore whether E2LLM can maintain its effectiveness with continued pretraining and whether long-context data can be synthesized from existing short-context data using LLMs. Furthermore, although E2LLM's performance is not highly sensitive to chunk size, selecting the optimal size can enhance its effectiveness, as shown in Table~\ref{table:chunk_size}. In the future, we aim to investigate more flexible and learnable chunking mechanisms.

\section*{Ethical Statements}
All the datasets (including benchmarks) and models used in this paper are publicly available resources. We use them only for research purpose.

\bibliographystyle{colm2024_conference}
\bibliography{custom}

\appendix
\clearpage

\section{Related Works}
\label{app:related_works}
As aforementioned, prevalent methods can be categorized into three groups: modifying the position embedding (i.e., length extension), the attention mechanism (i.e., sparse attention), and the input sequence (i.e., prompt compression).

\paragraph{Length Extension:} Training LLMs on sequences with limited maximum sequence lengths while ensuring generalization for longer sequences is challenging. To address this, positional extrapolation and interpolation methods have been proposed. Positional extrapolation extends positional encoding beyond the training length; for instance, ALiBi~\citep{presstrain} enhances attention with linear biases that adjust scores based on the distance between key and query positions. Instead, xPOS~\citep{sun2023length} utilizes relative position embeddings for better attention resolution and extended lengths. Another approach, CLEX~\citep{chen2024clex}, replaces manual design with learned scaling factors through neural differential equations, effectively overcoming the limitations inherent in traditional positional extrapolation techniques. Positional interpolation, on the other hand, scales down input position indices and expands context windows to maintain performance across longer sequences. For example, \citet{chen2023extending} applies linear interpolation to RoPE to align maximum position indices with pre-training constraints. NTK interpolation \citep{bloc97} modifies the base of RoPE to adjust the rotational velocity of its dimensions. To combine the strengths of these approaches, YaRN~\citep{peng2023YaRN} integrates linear and NTK interpolation with tailored extrapolation strategies, mitigating distribution shifts in the attention matrix with longer inputs. ResonanceRoPE~\citep{wang2024resonance} also targets reducing feature interpolation for out-of-distribution positions. LongRoPE~\citep{dinglongrope} further enhances performance by exploiting two forms of non-uniformities in RoPE positional embedding via an efficient evolutionary search. Besides modifying position embeddings, length extension can also be achieved by employing external memory for long contexts. CEPE \citep{yen2024long} adheres to the original Transformer architecture, using an encoder to process lengthy contexts chunk by chunk. The embeddings of tokens within each chunk given by the encoder are subsequently fed into the LLM through trainable cross-attention layers.

Despite these advancements, most approaches require continual pre-training or fine-tuning to achieve the desired length~\citep{zhao2024longskywork,gao2024train}, thus entailing a considerable training burden. Additionally, inference on these extended models can be slow due to the quadratic complexity of full attention. In contrast, the proposed E2LLM does not alter the original LLM's length but compresses the input sequence into chunks of embedding vectors. This allows E2LLM to maintain the efficiency of the original LLM during both training and inference.

\paragraph{Sparse Attention:} This category of methods aims to decrease the inference complexity of LLMs by manipulating attention mechanisms with novel attention masks, enabling these models to handle longer sequences. StreamingLLM~\citep{xiao2023efficient} demonstrates that focusing on the beginning of the sequence and the most recent tokens within a defined window (i.e., local attention) during inference maintains performance while significantly reducing computational costs to a linear scale. However, these training-free methods often fall short in various scenarios~\citep{anagnostidis2024dynamic,lou2024sparser}, as they may neglect informative tokens situated in the middle of the sequence. To improve performance, LM-Infinite~\citep{han2024lm} reintroduces top-k tokens from the middle, but this approach necessitates the computation of all attention scores, thereby increasing computational demands. As a solution, \citet{lou2024sparser} propose SparseK attention, which employs an additional scoring network to assess the importance of each key-value pair and select the top-k pairs. Alternatively, LongLoRA~\citep{chen2023extending} utilizes shifted sparse attention (a variant of local attention) and fine-tunes LLMs with LoRA~\citep{hu2021lora} to adapt to this mechanism. Unfortunately, as noted by~\citep{tan2024lloco}, there remains a significant gap between sparse and full attention, which complicates the fine-tuning of pre-trained LLMs to new attention paradigms. In contrast, the E2LLM approach summarizes long-context input into soft prompt vectors, thereby reducing context length without altering the full attention mechanism in LLMs.

\paragraph{Prompt Compression:} Prompt compression enhances the efficiency of LLM input processing by either condensing lengthy prompts (hard prompt compression) or learning compact prompt representations (soft prompt compression). Hard prompt compression techniques include RAG~\citep{ding2024survey}, LLMlingua~\citep{jiang2023llmlingua}, Selective-Context~\citep{li2023unlocking}, and LongLLMLingua~\citep{jiang2023LongLLMLingua}. RAG optimizes input by retrieving only the passages relevant to the query, while LLMlingua and Selective-Context focus on compressing extensive context without referencing the query. LongLLMLingua integrates these strategies by utilizing question-aware coarse-to-fine compression to enhance performance. However, these methods separate compression and inference into distinct steps, leading to potential error propagation that degrades performance. In contrast, E2LLM is trained end-to-end, effectively mitigating the above issue.

Soft prompt compression, proposed by \citet{mu2023learning} and \citet{ge2023context}, involves training LLMs to distill prompts into a more concise set of tokens that encapsulate the original prompt's knowledge for future use. \citet{chevalier2023adapting} extend this by developing AutoCompressor, which converts longer textual contexts into summary vectors that serve as soft prompts, which expands the LLM’s context window and reduces computational costs, as examplified in LLoCO~\citep{tan2024lloco}. However, directly using LLMs to generate sentence-level embeddings diverges from their original objective of next-token prediction. As a result, achieving satisfactory performance in this context often requires extensive training or fine-tuning to align the model with the new objective. To overcome this problem, our E2LLM leverages a pretrained sentence embedding model to represent prompts, aligning with the original training objectives of embedding models. Additionally, we note that, concurrently with our work, FocusLLM~\citep{li2024focusllm} has also adopted a strategy of chunking long contexts and summarizing each chunk using the hidden states of the local context from all layers of an LLM. These hidden states are concatenated to serve as the key-value cache for the same LLM, providing answers to user queries. From the perspective of E2LLM, FocusLLM essentially employs an LLM as a text encoder, which influences both training and inference efficiency.

\section{Overiew of Baseline Methods}
\label{app:baselines}

The following provides a brief overview of all baselines:
\begin{itemize}[leftmargin=*]
\item\textbf{Llama2-7B-chat}~\citep{touvron2023llama}: This refers to the original Llama2-7b-chat\footnote{\url{https://huggingface.co/meta-llama/Llama-2-7b-chat-hf}} without additional training or fine-tuning, serving as the backbone for the other methods.

\item\textbf{YaRN}~\citep{peng2023YaRN}: YaRN is a position interpolation method designed to effectively extend the context window of models trained with Rotary Position Embeddings (RoPE)~\citep{su2024roformer}. This method leverages the advantages of both linear and NTK interpolation. Note that the computational complexity of YaRN is quadratic in the context length during both training and inferece. We implement a scale factor of 16 and integrate LoRA~\citep{hu2021lora} into the self-attention module, utilizing a rank of 16. This results in a total of 17 million trainable parameters.

\item\textbf{CEPE}~\citep{yen2024long}: CEPE employs an encoder-decoder framework designed to efficiently manage long contexts by breaking them into manageable chunks. The encoder generates embeddings for each token within these chunks, which are then fed into the LLM decoder via cross-attention, in line with the original Transformer architecture. We use LLaMA-MLM-Large\footnote{\url{https://huggingface.co/hyen/LLaMA-MLM-Large}} as the encoder, with a total of 1.31B trainable parameters. During the warm-up stage, we train the cross-attention mechanism from scratch, followed by simultaneous training of both the encoder and cross-attention in the standard training phase. It is important to note that CEPE only presents a pretraining approach where the encoder initially processes a fixed-length segment of a sequence. This processed portion is then used to predict the remainder of the sequence for the decoder, functioning as a text completion task. In instances where a sequence is shorter than the predefined length of the encoder, the decoder is not provided with any input, which limits training flexibility. Unlike traditional Transformer fine-tuning, where the prompt and response are respectively inserted into the encoder and decoder, CEPE operates differently and does not support this method.

\item\textbf{StreamingLLM}~\citep{xiao2023efficient}: These approaches are training-free and utilize a $\Lambda$-shaped sparse attention mask, allowing tokens to only attend to the beginning of the sequence and recent tokens within a defined window. In our implementation of StreamingLLM, we set the start size at 4, while the recent size was set to 2000. 

\item\textbf{LongLoRA}~\citep{chen2023longlora}: This method utilizes shifted short attention instead of full attention during training and incorporates Position Interpolation~\citep{chen2023extending} and LoRA for fine-tuning an LLM to extend its context window. During inference, it reverts to full attention rather than sparse attention. We set the LoRA rank to 16 and fine-tune the self-attention, embeddings, and normalization modules, resulting in 140M trainable parameters.

\item\textbf{RAG}~\citep{gao2024retrievalaugmentedgenerationlargelanguage}: RAG operates with two core processes: retrieval and generation. During the retrieval phase, we adopt GTE-large-en~\citep{li2023generaltextembeddingsmultistage} as the retriever to recall the top-30 relevant context chunks, each with a maximum length of 512 characters, based on cosine similarity. These context chunks then serve as prompts for the LLM during the generation phase. Notably, RAG is training-free, offering flexibility in its application. However, it is essential to acknowledge that the retriever and the generator are distinct models trained on different corpora and with different objectives, which may lead to inconsistent interpretations of the same text~\citep{li2024unraveling, ding2024survey}.

\item\textbf{LongLLMLingua}~\citep{jiang2023LongLLMLingua}: This method builds upon the framework established by LLMLingua~\citep{jiang2023llmlingua} with the goal of identifying and removing non-essential tokens from prompts. This method begins by selecting passages, denoted as $x^\text{passage}$, that are relevant to the user query $x^\text{query}$ and that maximize the conditional probability $p(x^\text{passage}|x^\text{query})$. To achieve this, it utilizes an LLM, specifically the quantized Llama-7B-GPTQ, as a cross-encoder to rank the pairwise relevance of passages. It is important to note that cross-encoders tend to be significantly more computationally demanding than the bi-encoder retriever typically employed in RAG, although they offer higher accuracy. Once the relevant passages are identified, the method proceeds to select the most pertinent tokens $x_i$ from each passage, aiming to maximize the difference in perplexity: $\text{PPL}(x_i|x_{<i}) - \text{PPL}(x_i|x^\text{query},x_{<i})$. This process is also facilitated by the LLM. Ultimately, the selected tokens, limited to a total of 3000, are provided to the LLM to formulate an answer to the query. Note that the selected tokens may be non-consecutive, which can complicate the LLM's understanding of their semantic meaning.

\item\textbf{LLoCO}~\citep{tan2024lloco}: LLoCO utilizes Autocompressors~\citep{chevalier2023adapting} to encode long context offline into summary vectors or soft prompts. LLoCO omits the adapter used in E2LLM since its decoder is the same LLM (i.e., LLama2-7B-chat) as in the encoder AutoCompressor. As a result, the decoder can effectively understand the summary vectors generated by AutoCompressor after being fine-tuned with LoRA. One advantage of LLoCO is that its text encoder, AutoCompressor, considers the interdependencies of long-context chunks autoregressively. However, this also presents a limitation: the long context can only be processed sequentially, one chunk after another. By contrast, E2LLM can process all chunks in parallel and is more suitable for long context. Consistent with other methods, we employ LoRA on self-attention module with a rank of 16, resulting in the number of trainable parameters to be 17M. 
\end{itemize}

\section{Relation to Other Methods}
\label{sec:relation_others}
\subsection{Relation to Vision-Language Models}
E2LLM draws inspiration from recent advancements in Vision-Language Models (VLMs), including mini-GPT4~\citep{zhuminigpt}, LLaVA~\citep{liu2024visual}, Qwen-VL~\citep{wang2024qwen2}, and InternVL~\citep{chen2024internvl}. These VLMs utilize adapters to align pretrained vision encoders with LLM decoders, enabling the LLMs to process image tokens outputted by the encoders.
This approach leverages the strengths of independently pre-trained vision and language models, offering a flexible way to build high-performing systems.
Notably, \textbf{VLMs excel at performing OCR (Optical Character Recognition)}~\citep{islam2017survey}, effectively recognizing and outputting text present within images. 
Motivated by the success of VLMs, we propose that \textbf{by aligning text encoders with LLM decoders using an adapter, LLMs can similarly interpret encoded sentences and perform inference}. 
As both encoder and decoder operate within the text modality, this alignment process may be simpler and require less data than the cross-modal alignment used in VLMs. Furthermore, E2LLM's self-supervised reconstruction task allows us to leverage vast amounts of text to enhance the LLM's contextual understanding, unlike VLMs which rely on supervised image-text pairs that are more difficult to acquire.


\subsection{Relation to Retrieval-Augmented Generation}
\lzh{E2LLM shares a similar high-level structure with Retrieval-Augmented Generation (RAG), involving chunking, encoding, and decoding. However, the interaction between the encoder and decoder differs fundamentally. RAG retrieves relevant raw text passages using a pretrained encoder and feeds them directly into the decoder, relying entirely on textual inputs. In contrast, E2LLM compresses each chunk into a summary vector, allowing for more compact and semantically rich communication.}

\lzh{This design offers several advantages. First, E2LLM significantly extends context capacity by representing each chunk with a single vector token, enabling the decoder to handle thousands of chunks—far beyond the limitations of RAG’s fixed context window. Second, E2LLM aligns the encoder and decoder through an adapter module, reducing semantic inconsistencies often observed in RAG due to independent training. Lastly, E2LLM avoids reliance on sensitive retrieval hyperparameters by learning to reason over all chunks during training, making it more robust and generalizable. Notably, as shown in Section~\ref{ssec:rag}, E2LLM and RAG can complement each other, combining precise retrieval with scalable contextual reasoning.}

\subsection{Relation to Transformer}
\lzh{E2LLM departs from the standard Transformer architecture~\citep{c:22}, which was not originally designed to handle long-context reasoning. Unlike Transformers that pass all fine-grained tokens into the decoder, E2LLM compresses each chunk into a single chunk token via an encoder, significantly improving context scalability. Moreover, instead of relying on deep cross-attention layers between the encoder and decoder, E2LLM employs a lightweight adapter to align their representations. This design improves parameter efficiency while enabling effective information transfer, making E2LLM more suitable for long-context tasks with limited resources.}

\begin{table}[h]
\tabcolsep = 0.05cm
\begin{center}
\caption{Time and space complexity of various methods. $L$ and $C$ denote the context length and chunk size, respectively, while $M$ and $N$ respectively represent the number of initial starting tokens and recent tokens in StreamingLLM. Additionally, $\tau$ represents concurrency.}
\label{table:methods_time_space_complexity}
\resizebox{0.6\linewidth}{!}{
\renewcommand{\arraystretch}{1.}
\begin{tabular}{ccc}
\toprule
Methods       & Time Complexity                      & Space Complexity   \\ \midrule
Llama2-7B-chat     & $\gO(L^2)$                            & $\gO(L^2)$           \\
StreamingLLM  & $\gO(L(M+N))$                             & $\gO(L(M+N))$          \\
LongLoRA      & $\gO(L^2)$                             & $\gO(L^2)$           \\
CEPE          & $\gO(L(C/\tau+1/2)+L^2/4)$             & 
$\gO(L(C+1/2)+L^2/4)$ \\
YaRN          & $\gO(L^2)$                             & 
$\gO(L^2)$           \\
RAG           & $\gO(LC/\tau+C^2K^2)$ & $\gO(LC+C^2K^2)$   \\
LongLLMLingua & $\gO(L^2)$                             & $\gO(L^2)$           \\
LLoCO         & $\gO(LC+L^2/C^2)$                    & $\gO(LC+L^2/C^2)$  \\
E2LLM         & $\gO(LC/\tau+L^2/C^2)$             & $\gO(LC+L^2/C^2)$   \\ \bottomrule
\end{tabular}}
\end{center}
\end{table}

\section{Complexity of Existing Methods}
\label{app:complexity}
The original Llama2-7B-chat and YaRN rely on the quadratic time and space complexity inherent to the self-attention mechanism. In contrast, StreamingLLM modifies the attention strategy to focus solely on the initial $M$ starting tokens and $N$ recent tokens, resulting in a linear relationship between time and space complexity and the context length. Regarding LongLoRA, its inference process employs a global attention mechanism, leading to time and space requirements equivalent to those of YaRN and the original Llama2-7B-chat. CEPE divides the context into two segments, with the initial portion processed through parallelized embedding, represented in the table by the constant $\tau$ denoting concurrency, the subsequent self-attention and cross-attention mechanisms exhibit quadratic and linear complexities, respectively. RAG involves both the embedding and retrieval processes, establishing a direct correlation with the chunk size $C$ and the number of retrieved chunks $K$. An increase in $K$ results in slower speeds and greater space consumption, albeit with improved performance. For LongLLMLingua, it incorporates question-aware coarse-grained and fine-grained compression processes, which significantly consume time and space resources during the multiple computations of perplexity. LLoCO exhibit nearly identical time complexity to E2LLM, as both involve encoding and decoding processes. However, it is important to note that while E2LLM's encoding process shares similarities with the embedding process of RAG and can be executed concurrently, LLoCO is constrained by the AutoCompressor, which operates serially and thus cannot be parallelized. Moreover, the efficiency of both methods is directly tied to $C$, E2LLM benefits from high compatibility and can utilize long-context sentence embedding models such as BGE, GTE, and Jina-embedding as encoders, while LLoCO is limited by the AutoCompressor, restricting the chunk size range to 0-1536.

\begin{table}[h]
\tabcolsep = 0.02cm
\begin{center}
\caption{Dataset Statistics for Long Document Summarization and QA Tasks.}
\label{table:dataset_statistics}
\resizebox{0.6\linewidth}{!}{
\renewcommand{\arraystretch}{1.3}
\begin{tabular}{lcccc}
\toprule
Dataset     & Task Type     & \#Train. Samp. & \#Eval. Samp. & Len. \\ \midrule
QMSum       & Summarization & 1,257               & 272                & 14,428.78            \\
GovReport   & Summarization & 10,000               & 500                & 11,204.00           \\
Quality    & DocumentQA    & 5,046              & 2,086               & 6,797.66              \\
NarrativeQA & DocumentQA    & 3,000               & 200                & 52,158.88            \\
TriviaQA    & DocumentQA    & 10,000              & 500                & 1,075.90             \\ \bottomrule
\end{tabular}}
\end{center}
\end{table}

\section{More Details of Document Summarization and QA Tasks}
\label{app:doc_sum_qa}

\subsection{Datasets}
\label{app:doc_sum_qa_data}

In order to evaluate the effectiveness of E2LLM, we leverage five publicly available datasets that encompass both Summarization and Document Question-Answering (DocumentQA) tasks. The data statistics are shown in Table~\ref{table:dataset_statistics}.
\begin{itemize}[leftmargin=*]
\item{\textbf{QMSum}}\footnote{\url{https://github.com/Yale-LILY/QMSum}}~\citep{zhong2021qmsum} is a newly devised, human-annotated benchmark designed for the query-based multidomain meeting summarization task. It comprises an extensive range of query-summary pairs across 232 meetings in diverse fields. Specifically, we included 1,257 training samples and used 272 samples for inference. The average length of the samples in this dataset is 14,428.78 tokens.

\item{\textbf{GovReport}}\footnote{\url{https://huggingface.co/datasets/ccdv/govreport-summarization}}~\citep{huang2021efficient} contains elongated reports by the U.S. Government Accountability Offices and the Congressional Research Service, complemented by summaries and abstracts hand-written by experts, which is of the summarization task genre. For training purposes, 10,000 random samples were utilized, and for inference, 500 samples were arbitrarily selected from the validation sets. The average length of the sampled data is 11,204.00 tokens.

\item{\textbf{Quality}}\footnote{\url{https://huggingface.co/datasets/emozilla/quality}}~\citep{bowman2022quality} is a DocumentQA dataset comprising 5,046 training samples and 2,086 inference samples with contexts that have an average length of 6,797.66 tokens.
Further, we convert the original single-choice data format of the dataset into the QA format.

\item{\textbf{NarrativeQA}}\footnote{\url{https://github.com/google-deepmind/narrativeqa}}~\citep{narrativeqa} is another DocumentQA dataset, primarily extracted from comprehensive book texts and film scripts from varied sources. The challenge here lies in generating concise answers from potentially disordered and lengthier texts. We randomly sample 3,000 pieces of data for training, while randomly choosing 200 samples for inference. The average sample length is 52,158.88 tokens.

\item{\textbf{TriviaQA}}\footnote{\url{https://huggingface.co/datasets/mandarjoshi/trivia_qa}}~\citep{joshi2017triviaqa}is also a high-quality DocumentQA dataset that houses over 650K question-answer-evidence triples. It includes 95K question-answer pairings authored by trivia enthusiasts and independently sourced evidence documents. We selected 10,000 and 500 samples for training and inference respectively, with the average sample length amounting to 1,075.90 tokens.
\end{itemize}

\subsection{Metrics}
\label{app:doc_sum_qa_metrics}

For the task of Summarization, the performance of all methods is measured using the Rouge~\citep{lin2004rouge} metric, which operates by comparing the n-gram of the generated text with that of the reference text. Specifically, we leverage Rouge-1, Rouge-2, and Rouge-L to assess the overlap between the single-token, consecutive dual-tokens, and the longest common subsequence (LCS) in the generated text by LLM and the reference text. We also compute their geometric mean, denoted as G-mean, and higher values reflect higher quality of the generated summaries.

Concerning the task of DocumentQA, we adopt the method demonstrated by \citep{shaham2023zeroscrolls}, which computes the unigram overlap between the generated and reference answers. This is accomplished by normalizing white-spaces, lower-casing, excluding stopwords and punctuation. Based on the number of unigram tokens, in conjunction with the token quantity of the generated and reference answers, we calculate precision, recall, and F1. Again, a higher value indicates a more precise answer by the model.

\lzh{Overall, G-mean/F1 measures token-level or lexical overlap with the ground truth, while PPL evaluates semantic similarity, reflecting the generation quality from a semantic perspective. Therefore, we believe a comprehensive evaluation requires considering both G-mean/F1 and PPL.}

\begin{table*}[t]
\tabcolsep = 0.04cm
\begin{center}
\caption{Performance on Long-Context datasets. The best results are in \textbf{bold}, the second are \underline{underlined}, and the third are \uwave{wavy underlined}.}
\label{table:performance_more_metrics}
\resizebox{\linewidth}{!}{
\renewcommand{\arraystretch}{1.3}
\begin{tabular}{lccccccccccccccccccc}
\toprule
\multicolumn{1}{l}{{\multirow{2}{*}{Methods}}} &
  \multirow{2}{*}{\begin{tabular}[c]{@{}c@{}}Trainable\\ Parameters\end{tabular}} &
  \multirow{2}{*}{\begin{tabular}[c]{@{}c@{}}Context\\ Window\end{tabular}} &
  \multicolumn{4}{c}{QMSum} &
  \multicolumn{4}{c}{GovReport} &
  \multicolumn{3}{c}{Quality} &
  \multicolumn{3}{c}{NarrativeQA} &
  \multicolumn{3}{c}{TriviaQA} \\ \cline{4-20} 
\multicolumn{1}{c}{} &
   &
   &
  R1 &
  R2 &
  RL &
  G-mean &
  R1 &
  R2 &
  RL &
  G-mean &
  Prec. &
  Recall &
  F1 &
  Prec. &
  Recall &
  F1 &
  Prec. &
  Recall &
  F1 \\ \midrule
Llama2-7B-chat &
  0M &
  4K &
  21.90 &
  4.91 &
  14.21 &
  11.51 &
  10.68 &
  2.86 &
  5.46 &
  5.50 &
  6.16 &
  25.46 &
  9.38 &
  3.04 &
  13.52 &
  4.65 &
  6.72 &
  76.66 &
  12.06 \\
StreamingLLM &
  0M &
  4M &
  7.59 &
  1.15 &
  5.43 &
  3.62 &
  7.46 &
  3.39 &
  4.76 &
  4.51 &
  1.50 &
  5.50 &
  2.00 &
  OOM &
  OOM &
  OOM &
  8.43 &
  \uwave{76.99} &
  14.53 \\
LongLoRA &
  140M &
  100K &
  13.92 &
  4.82 &
  10.79 &
  8.98 &
  {\ul 27.04} &
  \textbf{9.92} &
  {\ul 16.29} &
  {\ul 16.35} &
  7.41 &
  9.99 &
  7.65 &
  OOM &
  OOM &
  OOM &
  13.03 &
  49.28 &
  19.69 \\
CEPE &
  1.31B &
  128k &
  19.22 &
  3.66 &
  {\ul 17.74} &
  10.77 &
  10.53 &
  1.08 &
  9.89 &
  4.82 &
  1.35 &
  {\ul 29.89} &
  2.33 &
  1.41 &
  \uwave{21.31} &
  2.19 &
  - &
  - &
  - \\
YaRN &
  17M &
  64K &
  21.54 &
  \uwave{5.34} &
  16.24 &
  \uwave{12.31} &
  12.93 &
  4.13 &
  5.69 &
  6.72 &
  \uwave{13.20} &
  19.42 &
  {\ul {13.80}} &
  OOM &
  OOM &
  OOM &
  \uwave{13.53} &
  49.45 &
  \uwave{20.22} \\
RAG &
  0M &
  $+\infty$ &
  \uwave{22.63} &
  5.26 &
  15.14 &
  12.17 &
  \uwave{24.92} &
  {\ul 8.56} &
  \uwave{14.15} &
  \uwave{14.45} &
  2.81 &
  \textbf{32.89} &
  4.85 &
  3.12 &
  {\ul {23.99}} &
  {\uwave{5.03}} &
  {6.23} &
  {\textbf{77.41}} &
  {11.15} \\
LongLLMLingua &
  0M &
  40K &
  16.42 &
  3.56 &
  12.18 &
  8.93 &
  8.63 &
  2.19 &
  5.20 &
  4.56 &
  9.13 &
  \uwave{26.34} &
  10.89 &
  \uwave{5.26} &
  \textbf{30.78} &
  4.53 &
  5.20 &
  {\ul{77.33}} &
  14.01 \\
LLoCO &
  17M &
  128K &
  {\ul 23.71} &
  {\ul 5.51} &
  \uwave{16.79} &
  {\ul 12.99} &
  11.69 &
  3.11 &
  5.18 &
  5.73 &
  \textbf{16.81} &
  15.03 &
  \textbf{14.37} &
  {\ul 11.85} &
  11.34 &
  {\ul 10.87} &
  \textbf{64.04} &
  64.03 &
  \textbf{63.21} \\ \midrule
E2LLM &
  11M &
  400K &
  \rbt{\textbf{25.92}} &
  \rbt{\textbf{6.70}} &
  \rbt{\textbf{21.34}} &
  \rbt{\textbf{15.47}} &
  \rbt{\textbf{29.14}} &
  \rbt{\uwave{7.94}} &
  \rbt{\textbf{27.08}} &
  \rbt{\textbf{18.43}} &
  \rbt{\ul{13.41}} &
  \rbt{15.32} &
  \rbt{\uwave{12.95}} &
  \rbt{\textbf{13.84}} &
  \rbt{13.61} &
  \rbt{\textbf{12.47}} &
  \rbt{{\ul 38.82}} &
  \rbt{39.43} &
  \rbt{{\ul 38.57}} \\ \bottomrule
\end{tabular}
}
\end{center}
\end{table*}

\subsection{Implementation Details}
\label{app:doc_sum_qa_details}
We employ the GTE-Large-en model\footnote{\url{https://huggingface.co/Alibaba-NLP/gte-large-en-v1.5}} as the encoder, and Llama2-7B-chat\footnote{\url{https://huggingface.co/meta-llama/Llama-2-7b-chat-hf}} as the decoder, with vPMA acting as the adapter. To ensure a fair comparison with CEPE and LLoCO, which both utilize Llama2-7B-chat as their decoder, we have fixed the decoder to be the same model. The output dimension of each attention head in the vPMA is set to 1024, and during chunking, the chunk size is set to 512. For our experiments, we split each dataset into training and validation sets with a 95:5 ratio, and perform training and evaluation independently for each dataset. Two training tasks are used for each dataset: the ``understandin'' task and the ``reasoning'' task. The weight for the ``understanding'' task is set as follows: 1e-7 for the QMSum dataset, 1e-7 for the GovReport dataset, 1e-9 for the Quality dataset, 1e-7 for the NarrativeQA dataset, and 1e-9 for the TriviaQA dataset. Both the GTE-Large-en and Llama2-7B-chat models are fine-tuned using the LoRA method, with ranks set to 16 and 8, respectively, and alpha values set to 16 and 8. The vPMA is fine-tuned fully. This configuration results in a total of 11M trainable parameters. Training is conducted on 16 A100 GPUs, with a batch size of 12, a learning rate of 1e-4, and a 100-step warm-up period. Early stopping is applied based on validation loss. We employ the Accelerate\citep{accelerate} library and the DeepSpeed distributed framework~\citep{rajbhandari2020zeromemoryoptimizationstraining} for training, and accelerate the training process using FlashAttention 2~\citep{Dao2023FlashAttention2FA} along with mixed-precision training techniques. During inference, all evaluations are set on a single A100 GPU with 80G memory, and for each method, we set do\_sample=False during generation and conduct a single run to evaluate the results.
\subsection{Detailed Discussion}
\label{app:doc_sum_qa_discuss}

The results for all baseline methods in Appendix~\ref{app:baselines} are presented in Table~\ref{table:performance_compare}. Following this, we now discuss the outcomes for each category of methods in detail. 

\paragraph{Soft prompt compression:} It is apparent that the proposed \textbf{E2LLM consistently achieves either the best performance or ranks within the top three across all nine evaluated methods}. The other soft prompt compression technique, LLoCO, also demonstrates commendable performance, especially in QA tasks, highlighting the effectiveness of soft prompt compression techniques. However, LLoCO's performance declines slightly in summarization tasks, which aligns with observations in its original publication (see Table 1 in~\citep{tan2024lloco}). LLoCO leverages AutoCompressor~\citep{chevalier2023adapting} as its text encoder, operating without additional training. AutoCompressor utilizes Llama2 to generate summary vectors for each chunk, designed to retain only the information necessary for subsequent chunks while discarding other potentially valuable content, as highlighted by~\citep{rau2024context}. In QA tasks, only the relevant portions of the long context are required to prompt the LLM for accurate answers, aligning well with AutoCompressor’s training objectives. In contrast, summarization tasks necessitate an overall understanding of the entire context. Consequently, since the summary vectors produced by AutoCompressor do not encapsulate all information within each chunk, LLoCO's performance in summarization is adversely affected. Unlike LLoCO, E2LLM directly leverages the chunk tokens for various tasks instead of predicting subsequent chunks.

\paragraph{Hard prompt compression:} Similar to LLoCO, the hard prompt compression method LongLLMLingua also excels in Document QA compared to summarization. The challenge of compressing long context into 3,000 non-consecutive tokens manifests in two significant ways: (i) the chosen token count is insufficient for summarizing the full long context; and (ii) the non-consecutiveness can hinder LLM comprehension, potentially leading to inaccurate answers. Additionally, the performance of this method is sensitive to hyperparameters, such as the chunk or passage size, which is crucial when selecting relevant passages for the query prior to token selection. These issues are also prevalent in RAG. Further complicating matters, the bi-encoder utilized in RAG may not retrieve relevant passages as effectively as the cross-encoder employed in LongLLMLingua. Inconsistencies can also arise when the retriever (encoder) and the generator (decoder) interpret the same text, as they are pretrained on different corpora~\citep{li2024unraveling,ding2024survey}. E2LLM addresses these issues by aligning the encoder and decoder through the adapter, which provides a global semantic embedding for each chunk and allows the decoder to utilize all chunks as inputs. This approach differs from selectively choosing some tokens from each chunk, enabling E2LLM to effectively retain relevant information and consistently surpass both hard prompt compression methods.

\begin{table*}[t]
\tabcolsep = 0.001cm
\begin{center}
\caption{Performance as a function of context length. The best results are in \textbf{bold}, the second are \underline{underlined}, and the third are \uwave{wavy underlined}.}
\label{table:context_bin}
\resizebox{\linewidth}{!}{
\renewcommand{\arraystretch}{1.2}
\begin{tabular}{l|cccccccccc|cccccccccc}
\toprule
Method & \multicolumn{10}{c}{QMSum} & \multicolumn{10}{|c}{NarrativeQA} \\ \midrule
Context Length & \multicolumn{2}{c}{0K-6K} & \multicolumn{2}{c}{6K-12K} & \multicolumn{2}{c}{12K-18K} & \multicolumn{2}{c}{18K-24K} & \multicolumn{2}{c}{24K+} & \multicolumn{2}{|c}{0-24K} & \multicolumn{2}{c}{24K-48K} & \multicolumn{2}{c}{48K-72K} & \multicolumn{2}{c}{72K-96K} & \multicolumn{2}{c}{96K+} \\
Metric & G-mean & PPL & G-mean & PPL & G-mean & PPL & G-mean & PPL & G-mean & PPL & F1 & PPL & F1 & PPL & F1 & PPL & F1 & PPL & F1 & PPL \\ \midrule
Llama2-7B-chat & 13.05 & 28.57 & 11.99 & 85.35 & 11.54 & 84.31 & \uwave{12.56} & 81.74 & {\ul 10.32} & 85.60 & 3.10 & 75.81 & {\ul 10.71} & 178.28 & \uwave{7.51} & 250.81 & 0.61 & 2303.08 & \uwave{2.48} & 2215.08 \\
StreamingLLM & 3.27 & 36.35 & 4.21 & 168.63 & 3.32 & 224.24 & 3.26 & 356.17 & 2.45 & 362.41 & 4.36 & 79.34 & 2.53 & 135.71 & OOM & OOM & OOM & OOM & OOM & OOM \\
LongLoRA & 5.91 & \uwave{12.92} & 8.13 & {\ul 13.17} & 8.30 & \uwave{14.65} & 9.66 & {\ul 15.97} & 7.44 & \uwave{17.31} & 3.23 & {\ul 11.93} & 9.47 & \textbf{12.17} & OOM & OOM & OOM & OOM & OOM & OOM \\
CEPE & 11.66 & 128.01 & 10.42 & 144.34 & 9.29 & 161.28 & 8.21 & 145.54 & 6.56 & 234.24 & 3.37 & 3568.12 & 2.65 & 2272.04 & OOM & OOM & OOM & OOM & OOM & OOM \\
YaRN & \uwave{13.57} & 14.52 & \uwave{12.10} & 14.02 & 12.88 & 17.06 & 11.49 & 17.75 & 6.33 & 18.90 & 7.19 & 13.94 & 6.59 & 17.16 & OOM & OOM & OOM & OOM & OOM & OOM \\
RAG & \rbt{\textbf{10.28}} & \rbt{\uwave{12.56}} & \rbt{14.65} & \rbt{\textbf{12.31}} & \rbt{\textbf{16.18}} & \rbt{\ul {14.55}} & \rbt{\ul {15.12}} & \rbt{\uwave{15.99}} & \rbt{10.00} & \rbt{{\ul 14.08}} & \rbt{4.32} & \rbt{\textbf{10.38}} & \rbt{4.28} & \rbt{33.08} & \rbt{4.64} & \rbt{47.79} & \rbt{3.85} & \rbt{46.55} & \rbt{2.77} & \rbt{44.80} \\
LongLLMLingua & 7.73 & \textbf{11.25} & 9.83 & 15.12 & 8.72 & 16.25 & 9.08 & 19.66 & 8.87 & 21.55 & \uwave{7.84} & 26.52 & 6.23 & 29.45 & 3.16 & \uwave{29.96} & 1.72 & \uwave{38.53} & 1.03 & \uwave{48.53} \\
LLoCO & {\ul 13.63} & 34.56 & {\ul 12.78} & 41.27 & \uwave{13.15} & 47.45 & 12.13 & 47.87 & \uwave{10.03} & 56.30 & {\ul 10.89} & \uwave{13.32} & \uwave{10.67} & \uwave{15.67} & {\ul 10.88} & {\ul 17.31} & {\ul 11.42} & {\ul 16.19} & {\ul 9.43} & {\ul 18.54} \\
E2LLM & \rbt{\textbf{15.74}} & \rbt{13.71} & \rbt{\textbf{16.02}} & \rbt{\uwave{13.25}} & \rbt{{\ul 15.23}} & \rbt{\textbf{14.01}} & \rbt{\textbf{15.20}} & \rbt{\textbf{13.77}} & \rbt{\textbf{15.61}} & \rbt{\textbf{13.42}} & \rbt{\textbf{12.08}} & \rbt{13.66} & \rbt{\textbf{12.52}} & \rbt{{\ul 12.91}} & \rbt{\textbf{12.65}} & \rbt{\textbf{13.79}} & \rbt{\textbf{12.43}} & \rbt{\textbf{13.35}} & \rbt{\textbf{12.01}} & \rbt{\textbf{14.29}} \\ \bottomrule
\end{tabular}}
\end{center}
\end{table*}

\begin{table}[t]
\tabcolsep = 0.1cm
\begin{center}
\caption{\lzh{Performance as a function of context length for Code Completion Task.}}
\label{table:context_bin_code_tasks}
\resizebox{0.5\linewidth}{!}{
\renewcommand{\arraystretch}{1}
\begin{tabular}{lccccc}
\toprule
Group     & 0K-3K & 3K-6K & 6K-9K & 9K-12K & 12K+  \\ \midrule
LCC       & 15.60 & 16.47 & 14.16 & 13.33  & 16.64 \\
RepoBench & 13.90 & 15.92 & 14.62 & 14.69  & 17.18 \\ \bottomrule
\end{tabular}}
\end{center}
\end{table}

\paragraph{Sparse attention:} On the flip side, the sparse attention method LongLoRA shows superior performance on summarization tasks but struggles with QA tasks. This disparity can be attributed to the shift shot attention mechanism utilized in LongLoRA, which allows for overlapping attention blocks and enhances global information flow—an essential aspect of summarization requiring a holistic view of all tokens. Nevertheless, the sparse attention mask limits information flow between two arbitrary tokens. Consequently, when relevant parts of the long context are inaccessible during Document QA, LongLoRA may fail to deliver accurate answers due to the loss of vital contextual information.  StreamingLLM is training-free and implements a $\Lambda$-shaped attention mask that further limits overall information flow. Without training, models initially designed with full attention struggle to adapt to this mask, diminishing their performance across all datasets. E2LLM addresses these challenges by employing the original full attention mask rather than resorting to sparse attention while effectively compressing passages into soft prompts (i.e., semantic summaries). This strategy enables E2LLM to consistently achieve superior performance compared to sparse attention methods.

\paragraph{Length extension:} Lastly, we observe that the length extension method, YaRN, strikes a balance between QA and summarization, generally finishing third best across all tasks and metrics. Like E2LLM, it encompasses all relevant information; however, as noted in previous research~\citep{chen2023extending}, attention mechanisms can become dispersed in exceedingly long contexts, diffusing focus across numerous token positions and achieving performance inferior to E2LLM. CEPE faces a similar challenge. Moreover, training the cross-attention layers in CEPE usually requires a vast amount of data (around 20 billion tokens, as suggested in~\citep{yen2024long}). This need arises because these layers are absent from the original language model (LLM). In our experiments, the number of tokens for each task is less than 0.1 billion, raising concerns that the cross-attention layers may not be sufficiently trained. Thus, integrating cross-attention layers into existing LLMs may pose compatibility issues without access to a substantial dataset for re-training. Additionally, CEPE operates within a pretraining framework in which the encoder processes a fixed-length segment of the sequence initially. This segment is then used to predict the remainder of the sequence for the decoder, effectively functioning as a text completion task. Notably, for TriviaQA, the context length is often shorter than the encoder's predefined length, leaving the decoder without any input. This results in the decoder producing irrelevant answers after training on the TriviaQA data. In contrast, E2LLM addresses the issue of attention dispersion encountered by length extension methods by not extending the decoder's length. Instead, it trains the decoder to interpret the soft prompts generated by the encoder, thereby enhancing performance.

\lzh{Moreover, we observe that the F1 score of E2LLM on the TriviaQA dataset is not particularly high. Our analysis suggests that this is mainly due to the limitations of the F1 evaluation metric. In certain cases, the answers generated by E2LLM are semantically relevant to the ground-truth but lack token-level overlap, which leads to a reduced F1 score. Representative case studies are presented in Section~\ref{sssec:case_studies_triviaqa}. Therefore, we argue that a comprehensive evaluation requires considering both G-mean/F1 and PPL. When both are considered, E2LLM demonstrates overall superior performance across all datasets.}
\subsubsection{Case Studies for TriviaQA}
\label{sssec:case_studies_triviaqa}
\lzh{We present four bad cases, each providing the context, query, ground-truth answer, E2LLM's prediction and the evaluation result w.r.t. F1 and PPL. The specific results are as follows:\\}

\noindent
\lzh{\textbf{Case \#1:}\\
\textit{Context:} Gremlins is a 1984 American comedy horror film directed by Joe Dante and released by Warner Bros....The owner Mr. Wing refuses to sell the creature to Randall, though his grandson secretly sells the mogwai to Randall but tells him to remember three important rules that must never be broken \textbf{do not expose the mogwai to bright lights or sunlight which will kill it, do not let it get wet, and never feed it after midnight.}...In the aftermath of the Gremlins rampage, Mr. Wing appears to collect Gizmo, scolding the Peltzers for their carelessness, thinking the Western world is not ready but comments that Billy might some day be ready to care for Gizmo properly. \\
\textit{Query:} According to the film Gremlins, if you are looking after a Gremlin you should not feed him after midnight, not get him wet and not expose him to what?\\
\textit{Ground-Truth:} bright light\\
\textit{Prediction:} The answer is Sunlight\\
\textit{F1:} 0\\
\textit{PPL:} 1.01\\}

\noindent
\lzh{\textbf{Case \#2:}\\
\textit{Context:} The molars or molar teeth are large , flat teeth at the back of the mouth....\textbf{The third , rearmost molar in each group is called a wisdom tooth. It is the last tooth to appear, breaking through the front of the gum at about the age of twenty,} although this varies from individual to individual....For example, the dentition of the Early Cretaceous monotreme Steropodon is similar to those of Peramus and dryolestoids, which suggests that monotremes are related to some pre-tribosphenic therian mammals, but, on the other hand, the status of neither of these two groups is well-established. \\
\textit{Query:} Which teeth are the third molars, and are always the last to erupt?\\
\textit{Ground-Truth:} maxillary third molar\\
\textit{Prediction:} wisdom tooth\\
\textit{F1:} 0\\
\textit{PPL:} 1.76\\}

\noindent
\lzh{\textbf{Case \#3:}\\
\textit{Context:} John Winston Ono Lennon, (born John Winston Lennon; 9 October 19408 December 1980) was an English singer and songwriter who co-founded the Beatles, the most commercially successful band in the history of popular music.... \textbf{He was murdered three weeks after its release.} Lennon revealed a rebellious nature and acerbic wit in his music, writing, drawings, on film and in interviews. \textbf{Controversial through his political and peace activism, he moved to Manhattan in 1971, where his criticism of the Vietnam War resulted in a lengthy attempt by Richard Nixon 's administration to deport him.}...In September 1980, Lennon commented about his family and his rebellious nature: He regularly visited his cousin, Stanley Parkes, who lived in Fleetwood.\\
\textit{Query:} In which city was John Lennon murdered?\\
\textit{Ground-Truth:} n y\\
\textit{Prediction:} New York\\
\textit{F1:} 0\\
\textit{PPL:} 1.23\\}

\noindent
\lzh{\textbf{Case \#4:}\\
\textit{Context:} \textbf{Marc Dutroux ( born 6 November 1956 ) is a Belgian serial killer and child molester,} convicted of having kidnapped, tortured and sexually abused six girls from 1995 to 1996, ranging in age from 8 to 19, four of whom he murdered....\textbf{Early life Born in Ixelles, Belgium,} on 6 November 1956 , Dutroux was the oldest of five children....\textbf{He owned seven small houses, most of them vacant, and used three of them for the torture of the girls he kidnapped.}...His wife was aware of all these activities. Second arrest In late 1995, Dutroux was arrested by police for involvement in a stolen luxury car racket.\\
\textit{Query:} Marc Dutroux hit the headlines over a 'house of horrors' in which country?\\
\textit{Ground-Truth:} Belgian\\
\textit{Prediction:} Belgium\\
\textit{F1:} 0\\
\textit{PPL:} 1.18\\}

\subsection{Performance v.s. Context Length}

Next, we investigate the sensitivity of the models' performance to variations in context length. \lzh{Specificallly, We select the widely used QMSum and NarrativeQA datasets to represent long document summarization and long document QA respectively, as their average lengths are longer than those of other datasets and are widely used by privious works~\citep{yen2024long,jiang2023LongLLMLingua,tan2024lloco}. Besides, we also select LCC~\citep{longcoder} and RepoBench~\citep{liu2023repobench} datasets to represent repository-level code completion task.} Then, we categorize samples from them into five groups based on their context lengths and then evaluate the perplexity (PPL) of the answers within each group. Our findings are summarized in Table~\ref{table:context_bin} and Table~\ref{table:context_bin_code_tasks}.

The results presented in the table indicate that E2LLM demonstrates a strong resilience to variations in context length for both summarization (QMSum) and question-answering (NarrativeQA) tasks, consistently achieving the best results among all models. This robustness can be attributed to the ``understanding'' task incorporated during the training of E2LLM (see Section~\ref{ssec:train_tasks}). By reconstructing different parts of the context, E2LLM effectively comprehends the information, regardless of its length. 
\lzh{Moreover, the strong resilience also benifits from E2LLM's balanced information compression methodology. Specifically, E2LLM compresses chunks of approximately equal lengths into chunk tokens, ensuring that the number of chunk tokens increases proportionally with the context length. This means that the \textbf{compression ratio remains consistent regardless of the context length, which contributes to its resilience against variations in context length.}}

Notably, the performances of YaRN, LongLoRA, CEPE, RAG, and LongLLMLingua also exhibit insensitivity to context length. On the other hand, LLoCO's performance declines slowly with increasing context length. Finally, streamingLLM and the original Llama2-7B-chat demonstrate sensitivity to context length; streamingLLM loses more information in the middle of the context as length increases due to its specific $\Lambda$-shaped attention mask, while Llama2-7B-chat struggles to handle long contexts altogether, as its maximum length has not been extended.

\begin{table*}[t]
\tabcolsep = 2pt 
\begin{center}
\caption{Data Statics for LongBench v2.}
\label{tab:long_bench_statics}
\resizebox{1\linewidth}{!}{
\renewcommand{\arraystretch}{1}
\begin{tabular}{lcccccccccc}
\toprule
Sub Domain & Domain & Lang. & Metric & Easy & Hard & Short & Medium & Long & Sample & Avg. Len \\ \midrule
Event ordering & Single-Document QA & EN & ACC & 8 & 12 & 1 & 15 & 4 & 20 & 176,880.40 \\
Detective & Single-Document QA & EN & ACC & 8 & 14 & 1 & 17 & 4 & 22 & 112,816.59 \\
Literary & Single-Document QA & EN & ACC & 11 & 19 & 6 & 17 & 7 & 30 & 186287.83 \\
Multi-news & Multi-Document QA & EN & ACC & 11 & 12 & 14 & 9 & 0 & 23 & 52,093.61 \\
Academic & Single\&Multi-Document QA & EN & ACC & 25 & 69 & 63 & 21 & 10 & 94 & 90,491.53 \\
Financial & Single\&Multi-Document QA & EN & ACC & 14 & 23 & 7 & 21 & 9 & 37 & 180,481.81 \\
Governmental & Single\&Multi-Document QA & EN & ACC & 9 & 32 & 15 & 16 & 10 & 41 & 168,729.10 \\
Legal & Single\&Multi-Document QA & EN & ACC & 12 & 21 & 25 & 5 & 3 & 33 & 64,818.30 \\
New language translation & Long In-context Learning & EN & ACC & 11 & 9 & 0 & 2 & 18 & 20 & 580,749.00 \\
Many-shot learning & Long In-context Learning & EN & ACC & 10 & 11 & 0 & 21 & 0 & 21 & 115,840.76 \\
User guide QA & Long In-context Learning & EN & ACC & 15 & 25 & 12 & 20 & 8 & 40 & 185,394.85 \\
Agent history QA & Long-dialogue History Understanding & EN & ACC & 9 & 11 & 20 & 0 & 0 & 20 & 33,724.85 \\
Dialogue history QA & Long-dialogue History Understanding & EN & ACC & 11 & 8 & 0 & 19 & 0 & 19 & 119,593.89 \\
Code repo QA & Code Repository Understanding & EN & ACC & 18 & 32 & 12 & 9 & 29 & 50 & 1071,100.42 \\
Table QA & Long Structured Data Understanding & EN & ACC & 8 & 10 & 4 & 10 & 4 & 18 & 543,243.06 \\
Knowledge graph reasoning & Long Structured Data Understanding & EN & ACC & 12 & 3 & 0 & 13 & 2 & 15 & 412,388.20 \\ \bottomrule
\end{tabular}
}
\end{center}
\end{table*}

\section{More Details of LongBench v2}
\label{app:lb_v2}
\subsection{Description of Longbench v2}
\label{ssec:lb_v2_description}
We utilize the LongBench v2 benchmark~\citep{bai2024longbench2} to evaluate the effectiveness of E2LLM. This benchmark comprises 503 challenging multiple-choice questions, characterized by diverse text lengths ranging from 8K to 2M words. It includes six major domains: single-document QA, multi-document QA, long in-context learning, long dialogue history understanding, code repository understanding, and long structured data understanding. The benchmark has been meticulously crafted by nearly 100 professionals from various backgrounds to rigorously assess models' long-context understanding capabilities. The entries are classified based on difficulty, with 192 categorized as “Easy” and 311 as “Hard.” They are also grouped by word count into three categories: “Short” (<32k words), “Medium” (32k-128k words), and “Long” (>128k words), containing 180, 215, and 108 samples, respectively. Additionally, the questions exhibit a balanced answer distribution, with options A, B, C, and D accounting for approximately 19\%, 25\%, 30\%, and 26\% of the total, respectively. Overall, LongBench v2's diverse sequence lengths, distributions, patterns, difficulties, and domains make it an excellent benchmark for assessing the capabilities of long-context models. Further details are provided in Table~\ref{tab:long_bench_statics}. Note that we utilize the Qwen2 tokenizer for token count statistics for each sample.

\subsection{Implementation Details}
\label{app:lb_imple}
We utilize GTE-Large-en\footnote{\url{https://huggingface.co/Alibaba-NLP/gte-large-en-v1.5}} as the encoder, Qwen2.5-7B-Instruct\footnote{\url{https://huggingface.co/Qwen/Qwen2.5-7B-Instruct}} as the decoder, and vPMA as the adapter. The output dimension of each attention head in vPMA is set to 896, and the chunk size during chunking is set to 512. We apply the SDFT method~\citep{yang-etal-2024-self} to generate a distilled dataset that closely matches the original distribution of the decoder, which is subsequently used for fine-tuning. The prompt for SDFT is as follows:
{\fontsize{9.5}{12} \selectfont
\begin{verbatim}
Below are an instruction that describes a 
task along with a reference answer. Using the 
reference answer as a guide, write your own 
response.
### Instruction:
Given the context: {context}
Please follow the instruction: 
Answer the question: {query}
### Reference Answer:
{answer}
### Response:
\end{verbatim}
}
The data generated through this method is used to train the model on two tasks: the ``understanding'' task and the ``reasoning'' task, with the weight for the ``understanding'' task set to 1e-9. GTE-Large-en is fine-tuned using the LoRA method (rank=32, alpha=32), and Qwen2.5-7B-Instruct is fine-tuned using LoRA with rank=16 and alpha=16. The vPMA adapter is fully fine-tuned. Training is performed on 32 A100 GPUs with a batch size of 12, a learning rate of 1e-4, and a 100-step warm-up period, with a total of 2.4K steps. Training is accelerated using the Accelerate library and the DeepSpeed distributed framework, along with FlashAttention 2 and mixed-precision training techniques. Additionally, we perform supervised fine-tuning (SFT) on the Qwen2.5-7B-Instruct model using the same dataset and 32 A100 GPUs.

For inference, all evaluations are carried out on a single 80GB A100 GPU to ensure a fair comparison. E2LLM inference is performed using the Transformers framework, while inference for the SFT Qwen2.5-7B-Instruct model, due to memory constraints, is carried out using the vLLM framework. We evaluate the model's performance and speed based on the sample category (e.g., ``Easy'', ``Short''), reporting the average values for both Accuracy and time consumption (in seconds) per sample. For each method, we set do\_sample=False during generation and conduct a single run to obtain the results. 

\subsection{Discussion on Generalizability}
\label{app:lb_generalizability}
\lzh{Regarding the generalizability of E2LLM, we believe it is validated by its competitive performance on the LongBench v2 benchmark. E2LLM is trained using a limited dataset comprising 13K samples from QMSum, GovReport, Quality, NarrativeQA, and TriviaQA (as explained in Section~\ref{ssec:lb_v2}). However, the scope of LongBench v2 is much broader than this training data and is disjoint from it (as discussed in~\ref{ssec:lb_v2_description}). Despite this, E2LLM still demonstrates strong performance, achieving results that surpass those of larger models such as Llama 3.1-70B and Nemotron-70B. Our goal in using the training data is to align the encoder and decoder in E2LLM and activate its long-context capabilities, rather than merely memorizing specific knowledge points. The results from LongBench v2 indicate that E2LLM can generalize effectively across a wide scope, even when trained on a relatively small amount of data.}

\lzh{To mitigate overfitting, we adopt the LoRA-based PEFT strategy instead of full fine-tuning, enabling efficient training of both the encoder and decoder with reduced overfitting risk. Additionally, we leverage the SDFT method, which guides the decoder to generate training samples based on reference answers rather than ground-truth labels, thereby preventing memorization. Early stopping is also employed to further enhance generalization.}

\begin{figure}[t]
    \centering
    \includegraphics[width=0.65\linewidth]{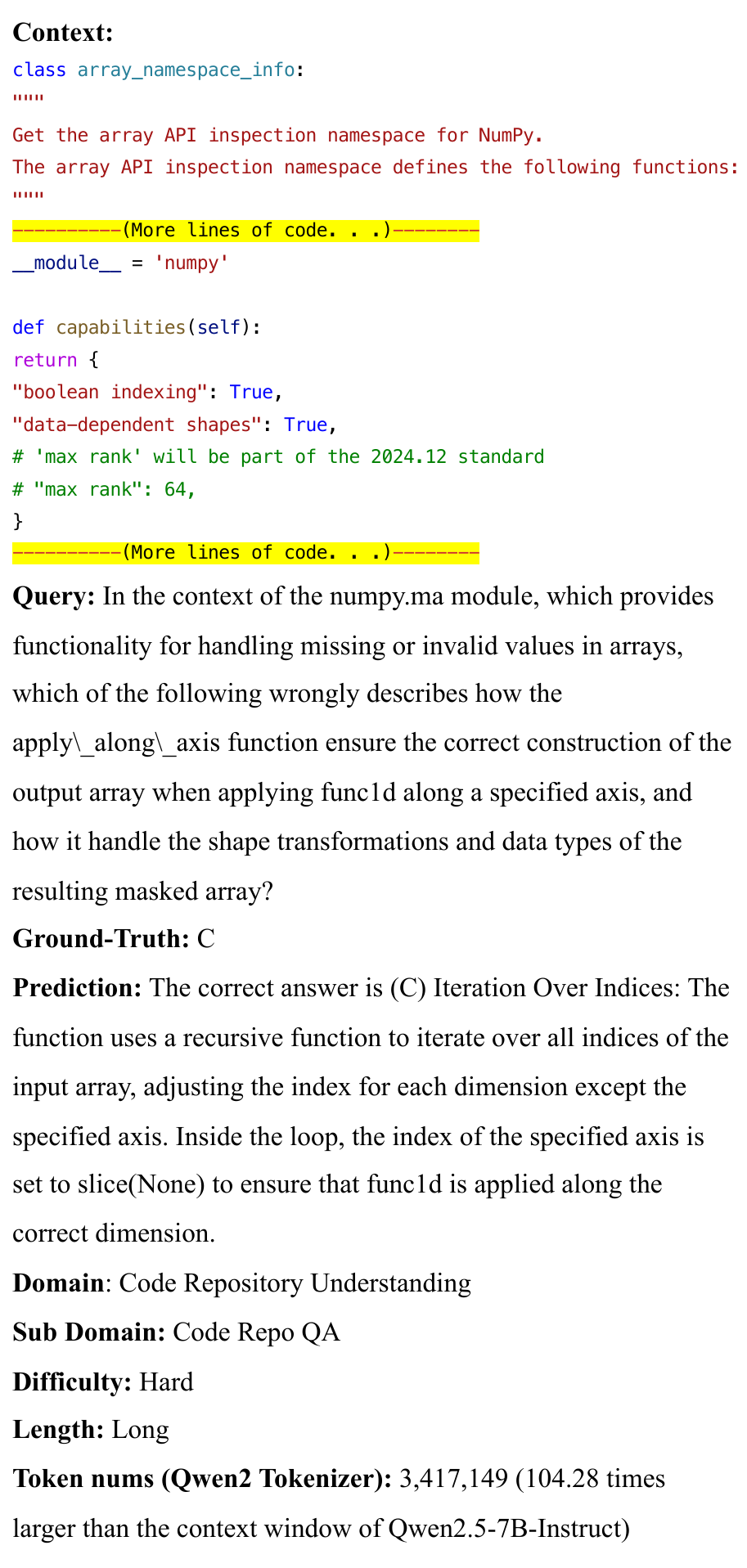}
    \caption{\lzh{Case study on extremely long-context input in LongBench v2.}}
    \label{fig:long-input}
    \vspace{-1ex}
\end{figure}
\subsection{Performance on Extremely Long Inputs}
\label{app:lb_extremely_long}
\lzh{To evaluate E2LLM's capabilities with extremely long-context inputs, we select test samples from LongBench v2 that exceed 95 times the original context length of the LLM decoder. The average token length of these samples is 3,508,190, compared to the original LLM decoder's maximum of 32,000 tokens. E2LLM achieved an accuracy of 50\% on these samples, significantly higher than the overall accuracy of 31.8\% (see Table~\ref{table:performance_longbench_v2}). This indicates that E2LLM effectively retains information when expanding the context length to nearly 100 times. One extremely long-context input case is represented in Figure~\ref{fig:long-input}.}

\begin{table*}[t]
\tabcolsep = 2pt 
\begin{center}
\caption{Task examples in RULER, the table is reproduced from~\citep{hsieh2404ruler}}
\label{tab:ruler_intro}
\resizebox{1\linewidth}{!}{
\renewcommand{\arraystretch}{1}
\begin{tabular}{ll}
\hline
\textbf{Task} & \textbf{Example} \\ \hline
\begin{tabular}[c]{@{}l@{}}Single\\ NIAH\\ (S-NIAH)\end{tabular} & \begin{tabular}[c]{@{}l@{}}(essays) ...... \\ One of the special magic numbers for long-context is: 12345. ...... \\ What is the special magic number for long-context mentioned in the provided text?\\ Answer: 12345\end{tabular} \\ \hline
\begin{tabular}[c]{@{}l@{}}Multi-keys\\ NIAH\\ (MK-NIAH)\end{tabular} & \begin{tabular}[c]{@{}l@{}}(essays) ...... \\ One of the special magic numbers for long-context is: 12345. \\ One of the special magic numbers for large-model is: 54321. \\ ...... \\ What is the special magic number for long-context mentioned in the provided text?\\ Answer: 12345\end{tabular} \\ \hline
\begin{tabular}[c]{@{}l@{}}Multi-values\\ NIAH\\ (MV-NIAH)\end{tabular} & \begin{tabular}[c]{@{}l@{}}(essays) ...... \\ One of the special magic numbers for long-context is: 12345. \\ One of the special magic numbers for long-context is: 54321. \\ ...... \\ What are all the special magic numbers for long-context mentioned in the provided text?\\ Answer: 12345 54321\end{tabular} \\ \hline
\begin{tabular}[c]{@{}l@{}}Multi-queries\\ NIAH\\ (MQ-NIAH)\end{tabular} & \begin{tabular}[c]{@{}l@{}}(essays) ...... \\ One of the special magic numbers for long-context is: 12345. \\ One of the special magic numbers for large-model is: 54321. \\ ...... \\ What are all the special magic numbers for long-context and large-model mentioned in the provided text?\\ Answer: 12345 54321\end{tabular} \\ \hline
\begin{tabular}[c]{@{}l@{}}Variable\\ Tracking\\ (VT)\end{tabular} & \begin{tabular}[c]{@{}l@{}}(noises) ...... \\ VAR X1 = 12345 ...... VAR Y1 = 54321 ...... \\ VAR X2 = X1 ...... VAR Y2 = Y1 ...... \\ VAR X3 = X2 ...... VAR Y3 = Y2 ...... \\ Find all variables that are assigned the value 12345. \\ Answer: X1 X2 X3\end{tabular} \\ \hline
\begin{tabular}[c]{@{}l@{}}Common Words\\ Extraction\\ (CWE)\end{tabular} & \begin{tabular}[c]{@{}l@{}}aaa bbb ccc aaa ddd eee ccc fff ggg hhh iii iii ......\\ What are the 10 most common words in the above list? \\ Answer: aaa ccc iii ......\end{tabular} \\ \hline
\begin{tabular}[c]{@{}l@{}}Frequent Words\\ Extraction\\ (FWE)\end{tabular} & \begin{tabular}[c]{@{}l@{}}aaa bbb ccc aaa ddd eee ccc fff ggg aaa hhh aaa ccc iii iii ......\\ What are the 3 most frequently appeared words in the above coded text? \\ Answer: aaa ccc iii\end{tabular} \\ \hline
\begin{tabular}[c]{@{}l@{}}Question\\ Answering\\ (QA)\end{tabular} & \begin{tabular}[c]{@{}l@{}}Document 1: ...... aaa ...... \\ Document 2: ...... bbb ...... \\ Document 3: ...... ccc ...... \\ Question: question \\ Answer: bbb\end{tabular} \\ \hline
\end{tabular}
}
\end{center}
\end{table*}

\section{Implementation Details for Augmentation with Retrieval}
We trained our E2LLM following specific settings for each task. For document summarization (Doc Sum) and question answering (Doc QA), we adhered to the configurations detailed in Appendix~\ref{app:doc_sum_qa}. For the LongBench v2 (LB v2) benchmark, we trained an E2LLM model using data generated through the SDFT method, as described in Section~\ref{ssec:lb_v2}, with implementation details found in Appendix~\ref{app:lb_imple}. The Needle-In-A-HayStack (NIAH) task, part of the RULER Benchmark, comprises eight distinct NIAH datasets. Training the E2LLM model for NIAH involved following the settings outlined in Appendix~\ref{app:ruler_imple}.


\section{RULER}
\label{app:ruler}

\subsection{Datasets}
In this section, we present the results of Llama2-7B-chat, LongLoRA, YaRN, LongLLMLingua, LLoCO, and E2LLM on the RULER benchmark~\citep{hsieh2404ruler}. RULER primarily consists of four types of tasks:
\begin{itemize} [leftmargin=*]
\item \textbf{Retrieval}: This task involves the Needle-in-a-Haystack test, which evaluates retrieval capability using diverse types and quantities of ``needles''. The term ``needle'' refers to a precise fact or statement concealed within a lengthy ``haystack'' of text. In RULER, NIAH encompasses various retrieval tasks, and the types of ``needles'' involved in these tasks include words, numbers, and essays.
\item \textbf{Muti-hop Tracing}: The variable tracking task (VT) serves as a minimal proxy for coreference chain resolution, examining the ability to trace entities across multi-hop connections. This task examines the ability to track relevant co-occurrence patterns and map the behavior of skipped connections within long context.
\item \textbf{Aggregation}: This task entails the extraction of common or frequent words (CWE and FWE), functioning as a proxy for summarization to test the ability to aggregate relevant information across long-range contexts. For CWE, words are sampled from a discrete uniform distribution. The model is tasked with identifying a fixed number of commonly used words from a set of infrequent words, the quantity of which increases as the sequence length grows.
For FWE, words are sampled from a Zeta distribution. In this case, the model’s objective is to identify the top-K most frequent words within the given context.
\item \textbf{Question Answering}: For this task, distracting information is added to the input of existing short-context QA datasets in order to assess question-answering capabilities at various context sizes.
\end{itemize}
The details of sample demonstrations for each task is presented in Table~\ref{tab:ruler_intro}.
For the VT task, we set the number of variable name-binding chains and the number of times binding variable names in each chain to be 1 and 4, respectively. For the CWE and FWE tasks, we set the frequency of ten common words to be 30, uncommon words to be 3, and alpha as 2.0. Finally, for the QA task, we use two single-hop short-context QA datasets SQuAD and HotPotQA.

\begin{table*}[t]
\tabcolsep = 0.2cm
\begin{center}
\caption{Performance on RULER Benchmark. The best results are in \textbf{bold}, the second are \underline{underlined}, and the third are \uwave{wavy underlined}.}
\label{tab:ruler_performance}
\resizebox{\linewidth}{!}{
\renewcommand{\arraystretch}{1.2}
\begin{tabular}{l|ccccc|ccccc|ccccc}
\toprule
Contex Length & \multicolumn{5}{c}{4K} & \multicolumn{5}{|c}{8K} & \multicolumn{5}{|c}{16K} \\ \midrule
Task & NIAH & VT & CWE & FWE & QA & NIAH & VT & CWE & FWE & QA & NIAH & VT & CWE & FWE & QA \\ \midrule
\rbt{LLama2-7B-chat} & \textbf{95.87} & \textbf{27.00} & \textbf{85.60} & \textbf{74.33} & \textbf{63.00} & - & - & - & - & - & - & - & - & - & - \\
LongLoRA & 41.07 & 1.60 & 16.60 & 9.33 & \uwave{55.50} & 38.19 & 2.20 & 13.40 & 10.33 & \textbf{44.00} & 36.78 & 2.00 & 5.80 & 4.00 & \textbf{52.00} \\
YaRN & 44.00 & {\ul 19.80} & 15.20 & 20.33 & {\ul 57.00} & \uwave{42.21} & 1.80 & 10.30 & 11.67 & \uwave{34.50} & \uwave{38.88} & 1.40 & 3.90 & 5.33 & 29.00 \\
LongLLMLingua & 27.44 & 5.20 & 7.60 & {\ul 44.67} & 14.50 & 30.44 & \uwave{4.20} & 5.70 & \textbf{24.33} & 16.0 & 32.75 & {\ul 7.00} & 2.00 & \textbf{27.33} & 15.50 \\
LLoCO & 0.44 & 0.00 & 27.70 & \uwave{24.67} & 32.50 & 0.19 & 0.00 & \textbf{24.10} & \uwave{17.00} & 28.50 & 0.41 & 0.00 & \textbf{20.90} & {\ul 22.67} & 20.00 \\
\rbt{E2LLM-C} & 2.45 & 1.20 & 6.80 & 13.67 & 35.50 & 3.50 & 0.60 & 6.30 & 11.67 & 31.00 & 2.06 & 1.60 & \uwave{7.30} & \uwave{10.33} & 28.50 \\
\rbt{E2LLM-R} & {\ul 78.28} & 3.00 & {\ul 41.50} & 0.00 & 35.00 & {\ul 59.50} & {\ul 4.60} & {\ul 21.50} & {\ul 19.33} & 33.50 & {\ul 56.00} & \uwave{5.40} & 6.40 & 2.33 & \uwave{40.00} \\
\rbt{E2LLM-C+R} & \uwave{66.22} & \uwave{7.20} & \uwave{32.40} & 2.00 & 37.00 & \textbf{60.66} & \textbf{8.60} & \uwave{21.10} & 12.00 & {\ul 35.50} & \textbf{57.78} & \textbf{10.20} & {\ul 9.70} & 3.00 & {\ul 41.00} \\ \midrule
Contex Length & \multicolumn{5}{c}{32K} & \multicolumn{5}{|c}{64K} & \multicolumn{5}{|c}{128K} \\ \midrule
Task & NIAH & VT & CWE & FWE & QA & NIAH & VT & CWE & FWE & QA & NIAH & VT & CWE & FWE & QA \\ \midrule
\rbt{LLama2-7B-chat} & - & - & - & - & - & - & - & - & - & - & - & - & - & - & - \\
LongLoRA & \uwave{34.75} & 0.40 & 1.80 & 1.67 & \uwave{33.50} & OOM & OOM & OOM & OOM & OOM & OOM & OOM & OOM & OOM & OOM \\
YaRN & 32.94 & 1.20 & 2.80 & 2.00 & 28.50 & OOM & OOM & OOM & OOM & OOM & OOM & OOM & OOM & OOM & OOM \\
LongLLMLingua & 26.06 & {\ul 6.20} & 0.30 & 11.33 & 18.50 & \uwave{23.25} & {\ul 5.20} & 0.30 & 13.33 & 15.0 & \uwave{20.88} & \textbf{5.20} & 0.40 & {\ul 21.67} & 4.50 \\
LLoCO & 0.25 & 0.00 & 0.10 & \textbf{24.00} & 4.50 & 0.31 & 0.00 & 2.40 & \uwave{15.67} & 9.00 & 0.28 & 0.00 & \textbf{3.30} & 4.33 & 2.00 \\
\rbt{E2LLM-C} & 2.36 & 0.80 & {\ul 5.50} & \uwave{12.67} & 28.00 & \textbf{2.63} & 0.60 & \textbf{4.90} & {\ul 26.33} & \uwave{29.50} & 2.60 & 1.00 & {\ul 2.50} & \uwave{8.67} & \uwave{7.50} \\
\rbt{E2LLM-R} & {\ul 48.13} & \uwave{5.60} & \uwave{5.40} & 1.67 & {\ul 34.00} & \textbf{50.13} & \uwave{4.40} & \uwave{2.60} & 10.67 & {\ul 31.00} & {\ul 40.67} & \uwave{4.40} & 1.80 & 6.67 & {\ul 23.50} \\
\rbt{E2LLM-C+R} & \textbf{52.66} & \textbf{8.60} & \textbf{6.70} & {\ul 15.00} & \textbf{40.50} & {\ul 49.34} & \textbf{6.20} & {\ul 3.20} & \textbf{32.67} & \textbf{34.50} & \textbf{48.38} & {\ul 5.00} & \uwave{2.30} & \textbf{26.67} & \textbf{27.50} \\ \bottomrule
\end{tabular}}
\end{center}
\end{table*}

\subsection{Implementation Details}
\label{app:ruler_imple}

For models requiring training, we combine the training data from Section~\ref{ssec:doc_sum_qa} (with details in Table~\ref{table:dataset_statistics}), resulting in around 13K samples. The dataset is divided into training and validation sets with a 95:5 ratio. All models are based on Llama2-7B-chat with configurations consistent with those in \red{Appendix~\ref{app:baselines}}. Specifically for E2LLM, we utilize GTE-Large-en~\footnote{\url{https://huggingface.co/Alibaba-NLP/gte-large-en-v1.5}} as the encoder and Llama2-7B-chat~\footnote{\url{https://huggingface.co/meta-llama/Llama-2-7b-chat-hf}} as the decoder, with the vPMA adapter integrated. Each attention head of the vPMA is configured with an output dimension of 1024, and during chunking, the chunk size is set to 512. We employ two training objectives, ``understanding'' and ``reasoning'', assigning a weight of 1e-9 to the ``understanding'' task. GTE-Large-en is fine-tuned using the LoRA method with a rank of 32 and alpha of 32, while Llama2-7B-chat is fine-tuned with LoRA at a rank of 16 and alpha of 16. The vPMA adapter undergoes full fine-tuning. The training process is carried out on 16 A100 GPUs, with a batch size of 12, learning rate of 1e-4, and a warm-up period of 100 steps. Early stopping is applied based on validation loss. The training is facilitated using the Accelerate library and DeepSpeed distributed framework, with FlashAttention 2 and mixed precision training techniques employed to expedite the process. 

For inference, we consider three variants of E2LLM: the original E2LLM that only receives chunk tokens as input (denoted E2LLM-C), E2LLM that only receives text retrieved by the text encoder (denoted E2LLM-R), and E2LLM that receives both the texts of the retrieved chunks and the chunk tokens of the unselected chunks (denoted E2LLM-C+R). It is important to note that we only select the top three chunks most relevant to the user query during retrieval. Additionally, we do not train E2LLM-R and E2LLM-C+R to adapt to the retrieved contexts. The inference process for E2LLM-C+R is prompted as follows: 
{\fontsize{9.5}{12} \selectfont
\begin{verbatim}
Given the context: [chunk tokens]
Some information that may be useful: 
[retrieved_text_tokens]
Please follow the instruction:
Answer the question: {query}
\end{verbatim}
}
All inference is conducted on a single A100 GPU with 80 GB of memory, and for each method, we set do\_sample=False during generation and conduct a single run to obtain the results.

\subsection{Discussions}

The results are summarized in Table~\ref{tab:ruler_performance}. Given the diverse tasks in RULER, we can clearly identify the strengths and weaknesses of each method. While YaRN and LongLoRA perform well in the QA task, they struggle significantly with the CWE and FWE tasks, likely due to attention distraction issues that hinder their focus on specific common or frequent words. Additionally, both methods face out-of-memory problems when the context length reaches or exceeds 64K, even when utilizing an A100 GPU with 80GB of memory, indicating that their space complexity is too high for resource-limited scenarios.

In contrast, LongLLMLingua excels in the FWE task but underperforms in others. Soft compression methods, such as E2LLM-C and LLoCO, offer a more balanced performance profile, achieving comparable results on the aggregation (CWE and FWE) and QA tasks. E2LLM-C tends to be more favorable for QA tasks, while LLoCO performs slightly better on aggregation. However, both E2LLM-C and LLoCO underperform on the NIAH task, which demands fine-grained, token-level retrieval of information. This limitation is inherent to the nature of soft prompt compression, which primarily retains semantic information while compressing away token-level details. For tasks like QA, where broader semantic understanding is sufficient to form a coherent response, E2LLM-C often achieves SOTA performance among the evaluated methods.

On the other hand, the performance on the NIAH task sees a significant improvement with the E2LLM-R+C configuration, which combines the original E2LLM with RAG. Indeed, E2LLM-R+C consistently achieves either the best performance or ranks within the top three among all evaluated methods across all tasks in RULER. Moreover, E2LLM-R+C generally outperforms E2LLM-R (RAG alone), indicating that the compressed chunk tokens retained by E2LLM-C still contain valuable information contributing to performance. Given that QA tasks are more commonly encountered in real-world applications than the synthetic NIAH task (as seen in LongBench v2), we argue that E2LLM remains a practical and efficient tool for long-context modeling. If fine-grained retrieval capabilities are crucial for a specific application, augmenting E2LLM with RAG offers an effective solution.

Lastly, we observe that all methods perform poorly on the VT task, which demands a nuanced understanding of long contexts, presenting a challenge that may be too great for existing baselines.

\section{More Discussions on Training and Inference Efficiency}
\label{app:efficiency}

\paragraph{Training Efficiency:} We assess the training throughput of all methods requiring training, including YaRN, LongLoRa, CEPE, LLoCO, and E2LLM. The experiments conducted on a single eight A100 GPU-equipped machine focus on measuring the number of processed tokens per second (tps), which serve as our evaluation metric. The configuration for all baselines adheres to the respective parameters specified in each of their original papers, and for our E2LLM, a chunk size of 512 characters is set. 

As demonstrated in Figure~\ref{fig:efficiency}(a), YaRN is clearly the least training-efficient method due to its necessary handling of the quadratic time complexity associated with the context length, stemming from its lack of original long context compression. LongLoRA, utilizing a sparse attention mechanism, offers slightly improved efficiency compared to YaRN by eliminating the need to compute the attention between some query-key pairs. Conversely, both CEPE and LLoCO demonstrate high throughput. CEPE initially processes all chunks of the long context in a parallel way, akin to E2LLM, but retains token-level embedding opposed to chunk-level embedding. This method then only trains the cross-attention linking the encoder and decoder, introducing linear time complexity relative to the long context length. In contrast, E2LLM trains the decoder relative to the compressed context length, thus explaining CEPE's higher throughput. Surpassing these, LLoCO performs remarkably well in training efficiency given that the summary vectors or soft prompt are prepared offline ahead of time, necessitating only the fine-tuning of the LLM decoder. E2LLM finally, processes context chunks in parallel during the encoding phase and fine-tunes the decoder module efficiently with LoRA, thus also demonstrating commendable training efficiency.

\paragraph{Inference Efficiency:} We now proceed to examine the inference efficiency of various methods. We begin by selecting seven differing context lengths that range from 1K to 73K; both YaRN and LongLoRA encounter out-of-memory issues at a context length of 74K. For each selected context length, we randomly select ten samples and truncate them to their predefined lengths. Upon averaging the runtime and GPU memory costs (i.e., peak allocated memory) over these samples, we reveal the results as a function of context length in Figure~\ref{fig:efficiency}(b) and \ref{fig:efficiency}(c).

Our model, E2LLM, exhibits the most impressive performance metrics, particularly in terms of runtime and memory usage, even for lengthy sequences of up to 73K tokens. In contrast, both YaRN and LongLoRA display significantly higher resource consumption, primarily due to the quadratic complexity inherent in full attention mechanisms during inference (notably, LongLoRA employs a full attention mask at this stage). Unlike LongLoRA, StreamingLLM utilizes a $\Lambda$-shaped sparse attention mask during inference, resulting in reduced time and memory costs. However, as indicated in the official implementation, for any given context, StreamingLLM must initially load the entire KV cache associated with that context. During the subsequent generation process, it utilizes Sink Attention to preserve the KV caches for both the starting and recent tokens. Consequently, in long-context scenarios, the memory usage and inference time for StreamingLLM still exhibit quadratic growth, \lzh{i.e., the KV cache cannot be compressed during the prefill stage~\citep{yang2024pyramidinfer}}.

On the other hand, CEPE demonstrates both time and space efficiency by computing cross-attention solely between the input to the decoder (such as a user query) and the encoder. This approach allows CEPE to achieve subquadratic complexity concerning long contexts. However, it focuses on token-level embeddings instead of chunk-level embeddings, which necessitates more time and memory compared to E2LLM.

Furthermore, LongLLMLingua modifies the LLM into a cross-encoder to identify the most relevant chunks and tokens related to the user query. Consequently, while its runtime increases dramatically with longer contexts due to the cross-encoder's high complexity, the memory usage remains stable. This is because the chunks can be processed sequentially, preventing significant memory overhead.

A similar trend is observed in another advanced prompt compression method, RAG. As we do not account for the memory costs associated with the retrieval process, and considering the retriever only recalls the 40 most relevant chunks from a lengthy context regardless of its total length, the generator's inference memory does not depend on context length. Nonetheless, since it processes the retrieved context token-by-token, the inference time and memory requirements still exceed those of E2LLM.

Lastly, LLoCO also enhances inference time through soft prompt compression; however, its text encoder, AutoCompressor, can only compress the original text by a maximum of 32 times, whereas E2LLM achieves an impressive compression factor of around 100 times. Furthermore, while AutoCompressor processes all chunks sequentially, E2LLM leverages parallel processing, further minimizing inference time.

\begin{table*}[t]
\tabcolsep = 3pt
\begin{center}
\caption{Ablation Study on QMSum and NarrativeQA. }
\label{table:ablation_study}
  \scalebox{0.8}{
\renewcommand{\arraystretch}{1.3}
\begin{tabular}{cccccccc|c}
\toprule
\multicolumn{1}{l}{\multirow{2}{*}{Variants}} & \multicolumn{4}{c}{QMSum} & \multicolumn{3}{c|}{NarrativeQA} & \multirow{2}{*}{\begin{tabular}[c]{@{}c@{}}Avg.\\ Rel. Diff.\end{tabular}} \\ \cline{2-8}
\multicolumn{1}{l}{} & R1 & R2 & RL & G-mean & Prec. & Recall & F1 &  \\ \midrule
E2LLM & \rbt{25.92} & \rbt{6.70} & \rbt{21.34} & \rbt{15.47} & \rbt{13.84} & \rbt{13.61} & \rbt{12.47} & - \\
-Und & \rbt{23.76} & \rbt{5.20} & \rbt{19.45} & \rbt{13.64} & \rbt{11.78} & \rbt{10.54} & \rbt{10.65} & \rbt{-14.78\%} \\
-$\gA$+$\text{MLP}$ & \rbt{25.37} & \rbt{6.55} & \rbt{18.75} & \rbt{14.61} & \rbt{13.53} & \rbt{13.79} & \rbt{12.35} & \rbt{-3.42\%} \\
-$\gE$ & \rbt{24.32} & \rbt{5.87} & \rbt{20.45} & \rbt{14.29} & \rbt{12.47} & \rbt{11.25} & \rbt{11.56} & \rbt{-9.27\%} \\
-$\gD$ & \rbt{23.98} & \rbt{5.23} & \rbt{20.22} & \rbt{13.64} & \rbt{12.23} & \rbt{11.13} & \rbt{11.49} & \rbt{-12.03\%} \\
+Overlap & \rbt{25.57} & \rbt{6.92} & \rbt{21.51} & \rbt{15.61} & \rbt{13.68} & \rbt{13.87} & \rbt{12.71} & \rbt{+0.90\%} \\
+BGE & \rbt{24.61} & \rbt{6.29} & \rbt{21.09} & \rbt{14.83} & \rbt{13.24} & \rbt{13.20} & \rbt{11.82} & \rbt{-4.15\%} \\
+Llama2-13B & \rbt{25.94} & \rbt{6.83} & \rbt{21.95} & \rbt{15.72} & \rbt{14.03} & \rbt{13.81} & \rbt{12.82} & \rbt{+1.73\%} \\ \bottomrule
\end{tabular}}
\end{center}
\end{table*}

\section{Results and Discussions on Ablation Studies}
\label{app:ablation}
In this subsection, we conduct ablation studies of E2LLM using the QMSum and NarrativeQA datasets, which serve as representative benchmarks for long-context summarization and document question-answering tasks, respectively. 
Details of each variant examined in Table~\ref{table:ablation_study} are outlined below.

\begin{itemize} [leftmargin=*]
\item\textbf{$-$Und} variant entails excluding the ``understanding'' task from our model and only employing the ``reasoning'' task for training purposes, which emphasis on the critical role that the ``understanding'' task plays within the model's performance.

\item\textbf{$-$$\mathbf{\mathcal{A}}$$+$\text{MLP}} \rbt{denotes the absence of the vPMA-based adapter, using the encoder's [CLS] token as the chunk embedding and an MLP for alignment with the decoder's input space. This serves as a baseline to assess the proposed adapter's effectiveness.}

\item\textbf{$-$$\mathbf{\mathcal{E}}$} denotes the freezing of encoder parameters, thereby allowing only the adapter and the decoder-only LLM to be trainable. This configuration aims to substantiate our hypothesis that a pretrained encoder alone is incapable of preserving the pertinent information that significantly impacts the performance of the LLM. Hence, maintaining the encoder's parameters as trainable is crucial.

\item\textbf{$-\mathbf{\mathcal{D}}$} entails keeping the decoder-only LLM frozen, in order to test whether the LLM can still adequately comprehend the output tokens from the adapter in the absence of any dedicated training.

\item\textbf{$+$Overlap} variant introduces an overlap of 30\% of the chunk size between sequential chunks during the chunking process. Moreover, within the scope of the ``understanding'' task's restatement operation, the model is required to restate the overlapping section of these chunks once.

\item\textbf{$+$BGE} variant test, on the other hand, involves replacing the GTE-large-en model with the BGE-m3 model as the encoder. This study seeks to affirm that our model maintains compatable with different sentence-embedding models serving as encoders.

\item\textbf{$+$Llama2$-$13B} configuration, similar in testing to the $+$BGE variant, is designed to verify the compatibility of our E2LLM with other LLMs serving as decoders. 
\end{itemize}

\begin{table*}[t]
\tabcolsep = 0.05cm
\begin{center}
\caption{Effect of chunk size on the model performance.}
\label{table:chunk_size}
\resizebox{1.\linewidth}{!}{
\renewcommand{\arraystretch}{1.3}
\begin{tabular}{ccccccccccccccccccc}
\toprule
\multirow{2}{*}{Chunk Size} &
  \multirow{2}{*}{\begin{tabular}[c]{@{}c@{}}Context\\ Window\end{tabular}} &
  \multicolumn{4}{c}{QMSum} &
  \multicolumn{4}{c}{GovReport} &
  \multicolumn{3}{c}{Quality} &
  \multicolumn{3}{c}{NarrativeQA} &
  \multicolumn{3}{c}{TriviaQA} \\ \cline{3-19} 
     &      & R1     & R2     & RL     & G-mean & R1     & R2     & RL     & G-mean & Prec.  & Recall & F1     & Prec.  & Recall & F1     & Prec.  & Recall & F1     \\ \midrule
128  & 100K & \rbt{25.13} & \rbt{6.22} & \rbt{20.65} & \rbt{14.78} & \rbt{28.20} & \rbt{6.74} & \rbt{25.84} & \rbt{16.99} & \rbt{13.04} & \rbt{14.68} & \rbt{12.34} & \rbt{13.36} & \rbt{13.14} & \rbt{12.13} & \rbt{34.95} & \rbt{36.90} & \rbt{33.28} \\
512  & 400K & \rbt{25.92} & \rbt{6.70} & \rbt{21.34} & \rbt{15.47} & \rbt{29.14} & \rbt{7.94} & \rbt{27.08} & \rbt{18.43} & \rbt{13.41} & \rbt{15.23} & \rbt{12.95} & \rbt{13.84} & \rbt{13.61} & \rbt{12.47} & \rbt{38.82} & \rbt{39.43} & \rbt{38.57} \\
1024 & 800K & \rbt{26.04} & \rbt{6.77} & \rbt{21.40} & \rbt{15.57} & \rbt{28.65} & \rbt{7.87} & \rbt{27.23} & \rbt{18.31} & \rbt{13.22} & \rbt{14.57} & \rbt{12.45} & \rbt{13.34} & \rbt{13.27} & \rbt{11.77} & \rbt{36.31} & \rbt{39.02} & \rbt{35.41} \\
2048 & 1.6M & \rbt{24.08} & \rbt{5.56} & \rbt{19.68} & \rbt{13.81} & \rbt{27.82} & \rbt{6.05} & \rbt{25.17} & \rbt{16.18} & \rbt{12.63} & \rbt{14.15} & \rbt{12.07} & \rbt{12.99} & \rbt{12.85} & \rbt{11.65} & \rbt{36.87} & \rbt{36.44} & \rbt{33.36} \\ \bottomrule
\end{tabular}}
\end{center}
\end{table*}

First, we assess the significance of the ``understanding'' task within E2LLM. Our findings indicate a substantial decrease in performance—by 16.39\%—when this task is omitted, highlighting its crucial role in helping E2LLM interpret the chunk embeddings produced by the encoder and further enhancing the performance of the ``reasoning'' task. 
\rbt{Next, we assess the importance of the proposed adapter, and conduct an evaluation by replacing it with the combination of the [CLS] token and MLP. The results showed a performance drop of 3.42\%, indicating that our adapter more effectively captures chunk information and aligns better with the decoder.}
Then, we examine the necessity of training the LoRA branches of the encoder and the decoder during alignment. As shown in Table~\ref{table:ablation_study}, the results for configurations -$\gE$ and -$\gD$ underscore the importance of training these components; without this training, E2LLM's performance diminishes by 9.08\% and 12.03\%, respectively. Finally, we explore the impact of replacing the chunker, text encoder, and LLM decoder within E2LLM (notated as +overlap, +BGE, and +Llama2-13B). Our analysis reveals that chunkers with overlapping segments (e.g., 30\% overlap) provide a modest performance boost. Additionally, employing more advanced encoders and decoders further enhances E2LLM's performance, suggesting that improvements in individual components can positively affect the overall system.

\begin{table}[t]
\tabcolsep = 0.06cm
\begin{center}
\caption{Performance on E2LLM with larger-scale model.}
\label{table:70b}
\resizebox{0.6\linewidth}{!}{
\renewcommand{\arraystretch}{1}
\begin{tabular}{lccccc}
\toprule
 & R1 & R2 & RL & G-mean & PPL \\ \midrule
E2LLM-Lm-7B & \rbt{25.92} & \rbt{6.70} & \rbt{21.34} & \rbt{15.47} & \rbt{13.66} \\
E2LLM-Lm-70B & \rbt{26.31} & \rbt{6.78} & \rbt{23.63} & \rbt{16.15} & \rbt{12.02} \\ \midrule
Rel. Improv. & \rbt{+1.50\%} & \rbt{+1.19\%} & \rbt{+10.73\%} & \rbt{+4.40\%} & \rbt{+12.01\%} \\ \bottomrule
\end{tabular}}
\end{center}
\end{table}

\subsection{Scaling to Larger-Scale Models}
\label{app:70b_exp}
We adopt Llama2-70B as the decoder to further validate the feasibility of E2LLM on larger-scale language models (denoted as E2LLM-Lm-70B). During training, we apply 4-bit quantization using QLoRA's Parameter-Efficient-Finetuning (PEFT) method. We conduct training and evaluating on QMSum, assessing its performance using the R1, R2, RL, G-mean, and PPL metrics as well as their relative improvement over those corresponding to the 7B model. The results are shown in Table~\ref{table:70b}.

As shown in the table, the performance of E2LLM significantly improves when using Llama2-70B, particularly in terms of Rouge-L and PPL. It is important to note that Rouge-1 and Rouge-2 evaluate unigram and bigram overlaps, respectively, measuring the match between the generated text and reference text at the word and phrase levels. In contrast, Rouge-L evaluates the similarity of the generated and reference texts based on the longest common subsequence (LCS), which measures structural similarity at the sentence level. This indicates that by leveraging a larger model, E2LLM is able to better capture the overall sentence structure and word order. Additionally, the reduction in PPL further demonstrates the model’s ability to generate more coherent and reasonable content.

\begin{figure}[!t]
    \centering
    \begin{minipage}[b]{0.24\linewidth}
    \centering
        \includegraphics[width=0.99\linewidth]{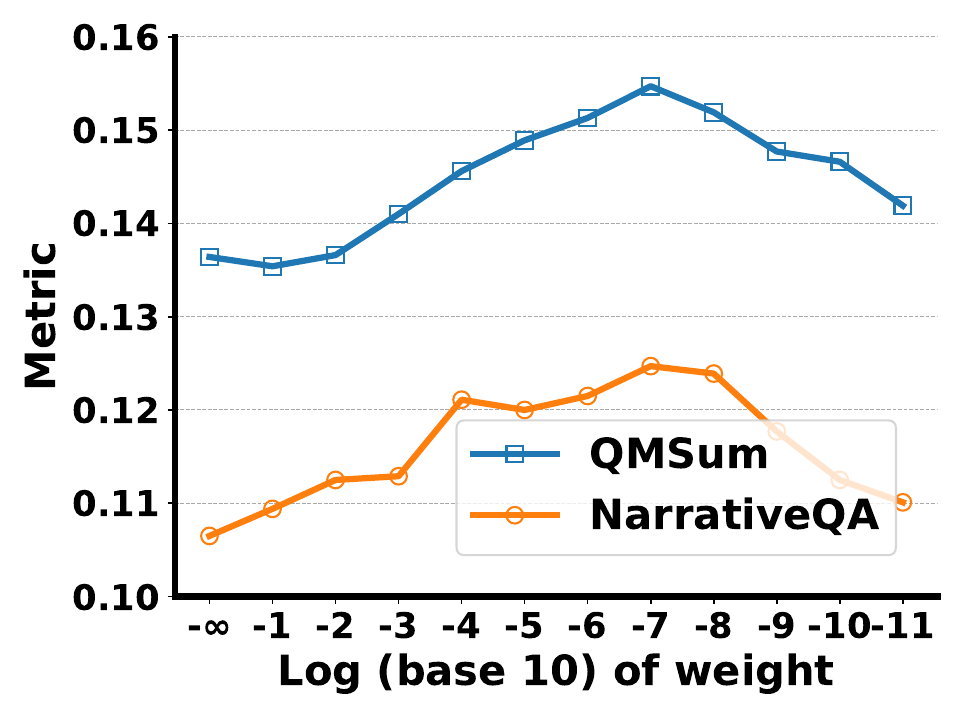}\\
    (a)
    \label{fig:param_weight}
    \end{minipage}
    \hfill
    \begin{minipage}[b]{0.24\linewidth}
    \centering
        \includegraphics[width=0.99\linewidth]{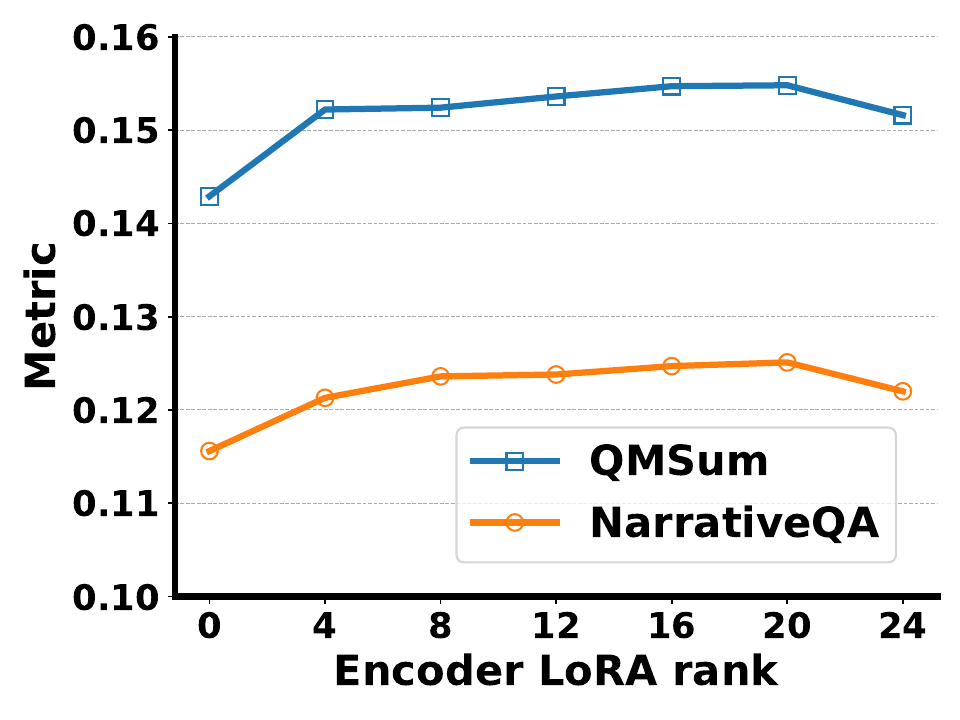}\\
    (b)
    \label{fig:param_mlp}
    \end{minipage}
    \hfill
    \begin{minipage}[b]{0.24\linewidth}
    \centering
        \includegraphics[width=0.99\linewidth]{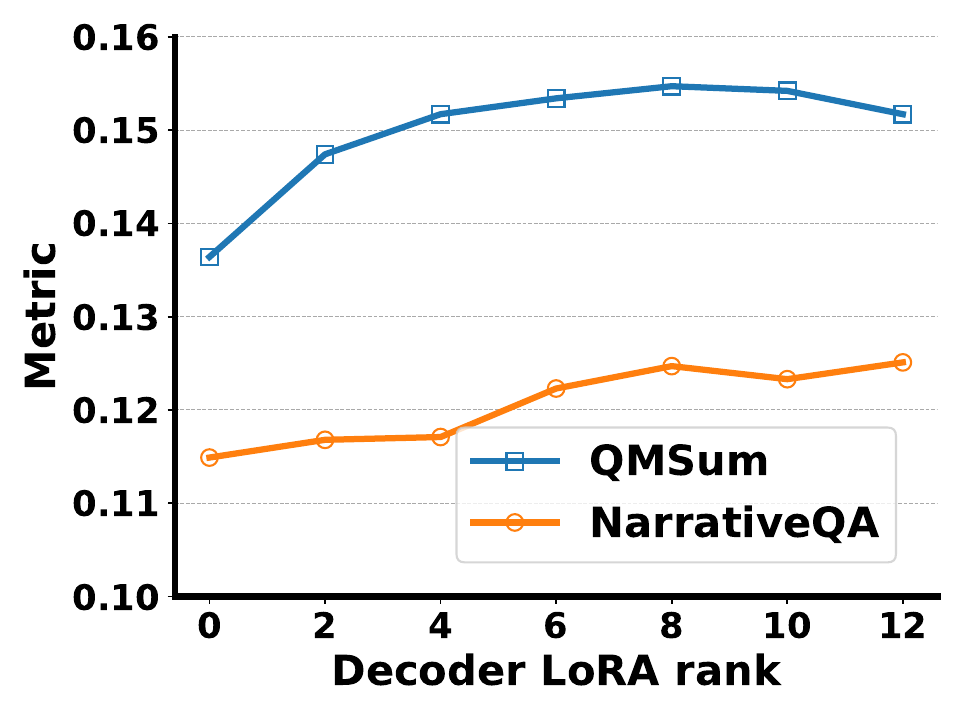}\\
    (c)
    \label{fig:param_encoder}
    \end{minipage}
    \hfill
    \begin{minipage}[b]{0.24\linewidth}
    \centering
        \includegraphics[width=0.99\linewidth]{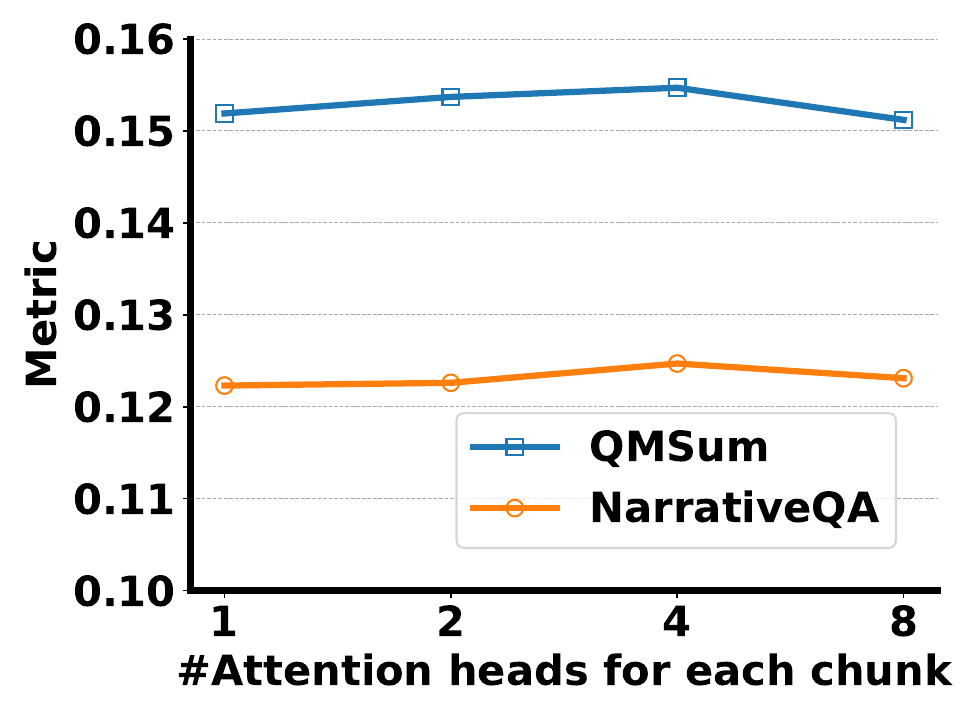}\\
    (d)
    \label{fig:param_attn}
    \end{minipage}
    \vspace{0.5ex}
    \caption{Effect of the hyperparameter. (a) the loss weight of ``understanding'' task. (b) the lora rank of encoder. (c) the lora rank of decoder. (d) \rbt{the number of attention heads for each chunk.}}
    \label{fig:param_investigation}
\end{figure}

\subsection{Hyperparameter Sensitivity}
\label{app:hyperparameter}

\begin{table}[t]
\tabcolsep = 0.03cm
\center
\caption{\lzh{Performance of E2LLM w.r.t. the weight assigned to the``understanding'' task.}}
\label{tab:augmented_weight}
\resizebox{0.6\linewidth}{!}{
\renewcommand{\arraystretch}{1}
\begin{tabular}{lcccccccccc}
\toprule
Dataset & \multicolumn{2}{c}{QMS.} & \multicolumn{2}{c}{G.R.} & \multicolumn{2}{c}{Qua.} & \multicolumn{2}{c}{N.QA} & \multicolumn{2}{c}{T.QA} \\ \hline
Metric & G-mean & PPL   & G-mean & PPL  & F1    & PPL  & F1    & PPL   & F1    & PPL  \\ \midrule
1e-7   & 15.47  & 13.66 & 18.43  & 2.81 & 12.36 & 8.45 & 12.47 & 13.07 & 33.85 & 7.90 \\
1e-9   & 14.77  & 13.98 & 17.49  & 2.97 & 12.95 & 8.99 & 11.77 & 12.93 & 38.57 & 7.53 \\ \bottomrule
\end{tabular}}
\end{table}

In this section, we explore the effects of hyperparameters on the performance of E2LLM, specifically focusing on the weight assigned to the ``understanding'' task, the LoRA rank of the encoder and decoder, the number of \rbt{attention heads in vPMA for each chunk}, and the chunk size.

The weight assigned to the ``understanding'' task indicates its relative importance compared to the ``reasoning'' task. Recall that the input context typically has a much longer length than answers, making it too long to be fully reconstructed at once. To address this, we employ a sliding window approach, reconstructing the original context in segments based on a few consecutive chunks until the entire input has been reconstructed. Consequently, the samples for the ``understanding'' task are significantly more numerous than those for the ``reasoning'' tasks. To maintain sample balance, we usually assign a smaller weight to the restatement task. As depicted in Figure~\ref{fig:param_investigation}(a), the optimal weight may vary across different datasets, which may be influenced by factors such as context length and the sentence embedding model's capacity to comprehend the specific semantics of the context. 
\lzh{Besides, we perform parameter sensitivity experiments with weights set to $1e-7$ and $1e-9$ on QMSum, GovReport, Quality, NarrativeQA, and TriviaQA. The results are summarized in Table~\ref{tab:augmented_weight}. It is easy to observe that while the weight for the ``understanding'' task does have some effect, its variation does not significantly impact the overall performance of E2LLM. Specifically, regarding the PPL, changing the weight between $1e-7$ and $1e-9$ does not affect E2LLM's ranking in Table~\ref{table:performance_compare} across all datasets, except for GovReport.}

Moreover, we investigate the optimal LoRA rank of the encoder (i.e., GTE-large-en) and the decoder (i.e., \rbt{Llama2-7B-chat}) within the range of \{0, 4, 8, 12, 16, 20, 24\} and \{0, 2, 4, 6, 8, 10, 12\}, respectively. The findings suggest that having no trainable parameters—in other words, completely ``freezing'' the encoder and decoder—hinders the effective extraction of original context content and alignment between the encoder and decoder, as discussed in Section~\ref{sec:model_arch}. As the rank of the two modules increases, a corresponding improvement in performance is observed, thereby underscoring the importance of training. Performance enhancement continues until it reaches a peak within a specific range of ranks. However, beyond this optimal range, further increases in rank lead to a decline in performance, attributable to overfitting on the training datasets. 

We also explore the influence of the number of attention heads within each chunk's vPMA on overall performance. We test the following values: {1, 2, 4, 8}. Our results, presented in Figure~\ref{fig:param_investigation}(d), consistently show that allocating 4 attention heads achieves superior performance across all datasets, demonstrating the stability of this configuration. We believe that fewer attention heads are insufficient to capture the various nuances in the relationships between token-level embeddings, leading to poorer results. Conversely, allocating a larger number of heads results in lower-dimensional representations for each head, which reduces their individual capacity to encode information, ultimately degrading performance.

We investigate the effect of chunk size on model performance, experimenting with sizes of {128, 512, 1024, 2048} characters, corresponding to maximum context window sizes of 100K, 400K, 800K, and 1.6M tokens for various E2LLM variants. Results in Table~\ref{table:chunk_size} show that the differences in performance metrics across different chunk sizes are relatively small for all datasets used in this study, indicating that the alignment process in E2LLM can effectively mitigate the impact of chunk size on performance. Nonetheless, selecting an optimal chunk size can still provide a slight performance boost. While smaller chunks might reduce compression and better preserve inputs, they may hinder context capture in longer sentences or paragraphs, making it difficult for the encoder to grasp semantics, which affects downstream tasks. Conversely, larger chunk sizes increase diversity and noise, complicating semantic capture and leading to decreased performance, especially in tasks like DocumentQA where relevant sentences may be overlooked. 

\end{document}